\documentclass[10pt,twocolumn,letterpaper]{article}

\usepackage{iccv}
\usepackage{times}
\usepackage{epsfig}
\usepackage{graphicx}
\usepackage{amsmath}
\usepackage{amssymb}
\usepackage{array}
\usepackage{booktabs}
\usepackage{tabularx}
\usepackage{multirow}
\usepackage{enumitem}
\usepackage[format=plain,labelformat=simple,labelsep=period,font=small,skip=4pt,compatibility=false]{caption}
\usepackage[font=scriptsize,skip=2pt]{subcaption}
\usepackage{mathtools}
\newcommand{\invisible}[1]{}
\newcommand{\myparagraph}[1]{\medskip\noindent\textbf{#1}}
\newcolumntype{B}[1]{>{\centering\arraybackslash}b{#1}}
\usepackage[pagebackref=true,breaklinks=true,letterpaper=true,colorlinks,bookmarks=false]{hyperref}

\usepackage{etoolbox}
\usepackage[binary-units]{siunitx}
\robustify\bfseries
\DeclareSIUnit{\inch}{inch}

\graphicspath{ {./Images/} }
\usepackage[capitalize]{cleveref}
\crefname{section}{Sec.}{Section}

\iccvfinalcopy % *** Uncomment this line for the final submission

\def\iccvPaperID{7} % *** Enter the ICCV Paper ID here

\ificcvfinal\fi

\begin{document}

%%%%%%%%% TITLE
\title{Joint Wasserstein Autoencoders for Aligning Multimodal Embeddings}

\author{Shweta Mahajan \qquad Teresa Botschen \qquad  Iryna Gurevych \qquad Stefan Roth\\[0.5em]
Department of Computer Science, TU Darmstadt\\
%{\tt\small www.\{aiphes,visinf,ukp\}@tu-darmstadt.de}
}

\maketitle
\thispagestyle{empty}

%%%%%%%%% ABSTRACT
\begin{abstract}
    One of the key challenges in learning joint embeddings of multiple modalities, e.g.~of images and text, is to ensure coherent cross-modal semantics that generalize across datasets.
    We propose to address this through joint Gaussian regularization of the latent representations.
    Building on Wasserstein autoencoders (WAEs) to encode the input in each domain, we enforce the latent embeddings to be similar to a Gaussian prior that is shared across the two domains, ensuring compatible continuity of the encoded semantic representations of images and texts.
    Semantic alignment is achieved through supervision from matching image-text pairs. 
    To show the benefits of our semi-supervised representation, we apply it to cross-modal retrieval and phrase localization.
    We not only achieve state-of-the-art accuracy, but significantly better generalization across datasets, owing to the semantic continuity of the latent space.
\end{abstract}

%%%%%%%%% BODY TEXT
\section{Introduction}

The availability of significant amounts of image-text data on the internet (\eg, images with their captions) has posed the question whether it is possible to leverage information from both visual \emph{and} textual sources.
To take advantage of such heterogeneous data, one of the fundamental challenges is the joint representation of multiple domains \cite{BaltrusaitisAM17}.
Powerful multimodal representations are integral to the accuracy of models on cross-domain tasks, such as image captioning \cite{KarpathyF17} or cross-domain retrieval \cite{AndrewABL13,EisenschtatW17,FaghriFKF17,KleinLSW15,LeeCHHH18,wang2018learning}.

Multimodal embeddings of images and texts can be obtained by mapping input image and text representations into a common latent space \cite{FaghriFKF17,wang2018learning}. 
Learning such representations is often formulated as an image-text matching problem in a fully supervised setup.  
An alternative approach is to first learn separate latent spaces and to align them later through constraints, \eg, supervised information \cite{TsaiHS17}.
The benefit is that the latent representations of each modality are learned independently, allowing to take advantage of unsupervised (\ie~unpaired) data.

\begin{figure*}[t!]
    \centering
    \begin{subfigure}{.35\textwidth}
        \centering
        \includegraphics[width=0.98\linewidth]{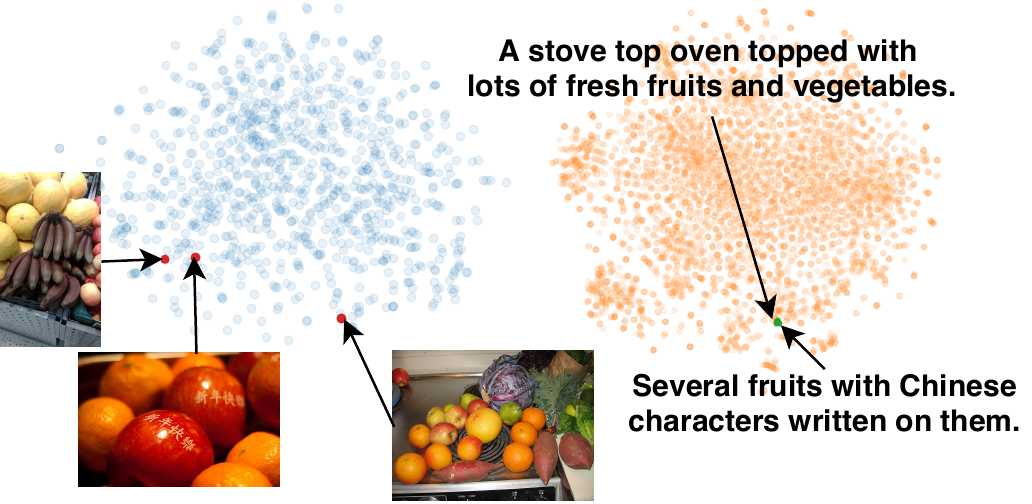}
        \caption{Input embeddings for images (left) and text (right).
        }
        \label{fig:teaser_a}
    \end{subfigure}%
    \hspace{0.35cm}
    \begin{subfigure}{.24\textwidth}
        \centering
        \includegraphics[width=0.9\textwidth]{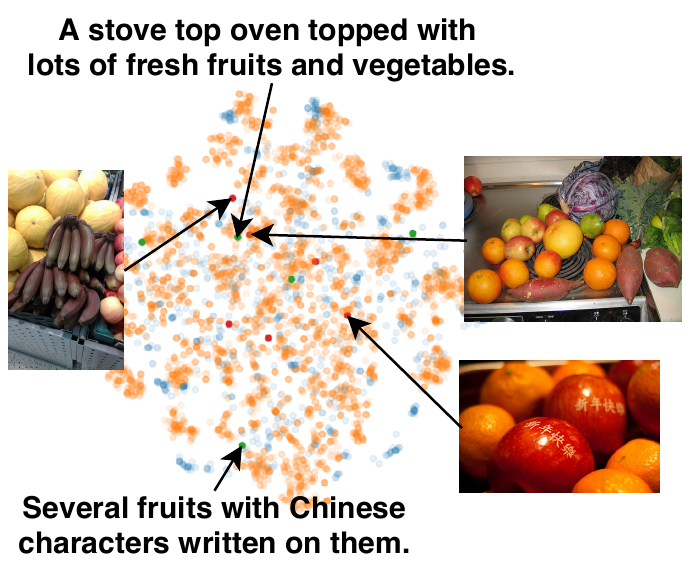}
        \caption{Joint representation of image-text embeddings based on negative sampling \cite{wang2018learning}.}
        \label{fig:teaser_b}
    \end{subfigure}%
    \hspace{0.35cm}
    \begin{subfigure}{.35\textwidth}
        \centering
        \includegraphics[width=\textwidth]{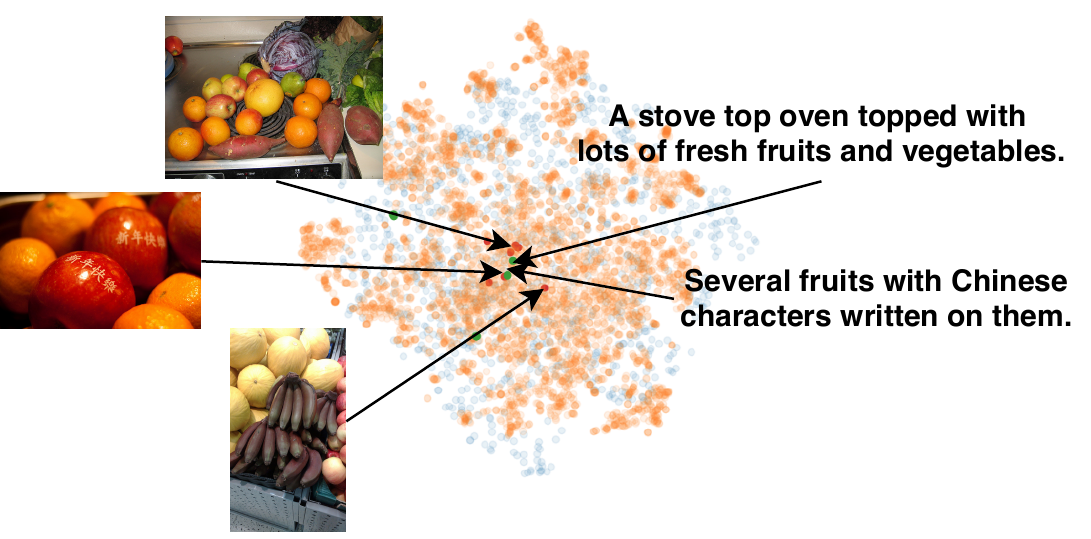}
        \caption{Coordinated representation of image-text embeddings of the proposed jWAE.}
        \label{fig:teaser_c}
    \end{subfigure}
    \caption{\emph{From input embeddings to coordinated embeddings with semantic continuity:}
        \emph{(\subref{fig:teaser_a})} Semantic similarity in the input modalities does not imply proximity in the embedding space.
        \emph{(\subref{fig:teaser_b})} Joint representations align the modalities, but the sparsity of supervised information does not achieve continuity of the semantic space.
        \emph{(\subref{fig:teaser_c})} Our jWAE based on joint Gaussian regularization leads to semantic continuity.
    }
\label{fig:teaser}
\end{figure*}

\begin{figure*}[t!]
    \centering
    \begin{subfigure}{.48\textwidth}
        \centering
        \includegraphics[width=\linewidth]{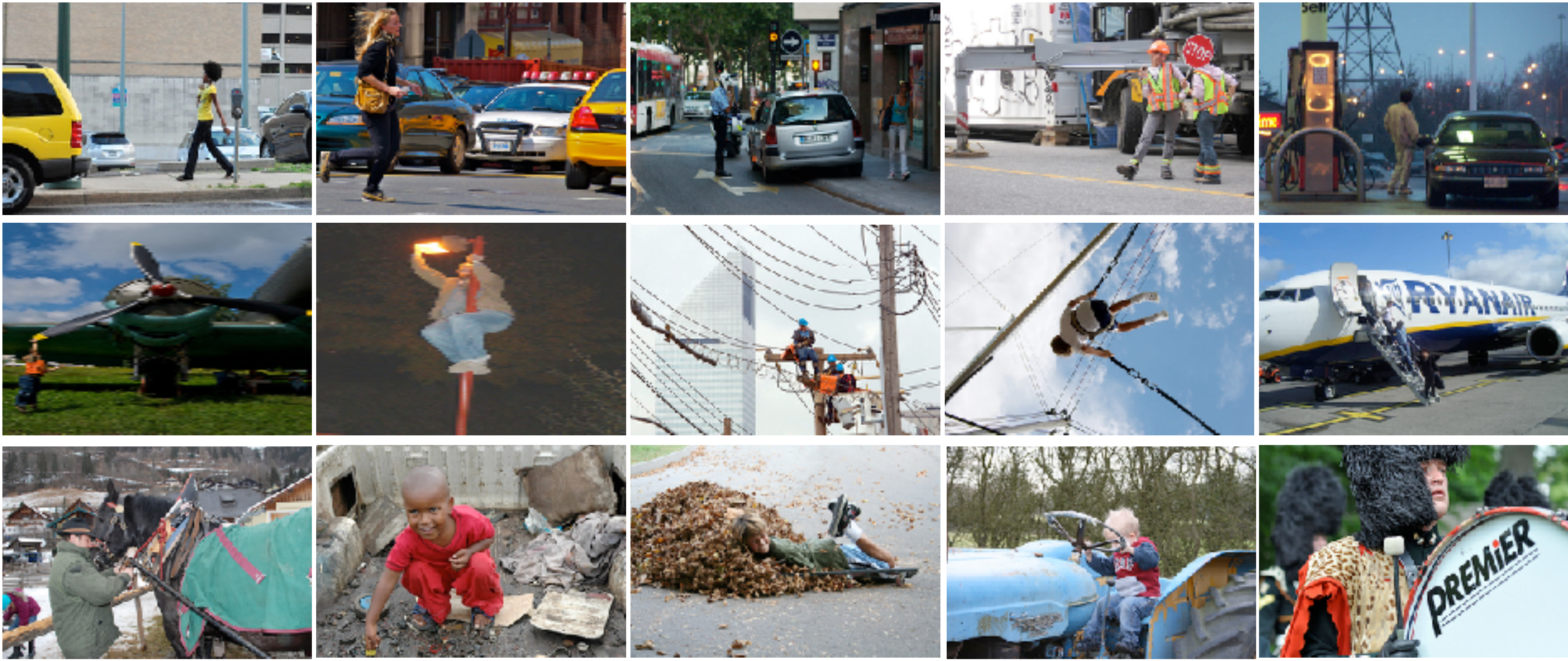}
        \caption{ Matching images in the joint embedding space of \cite{wang2018learning}.
        }
        \label{fig:semantic_a}
    \end{subfigure}%
    \hspace{0.5cm}
    \begin{subfigure}{.48\textwidth}
        \centering
        \includegraphics[width=\linewidth]{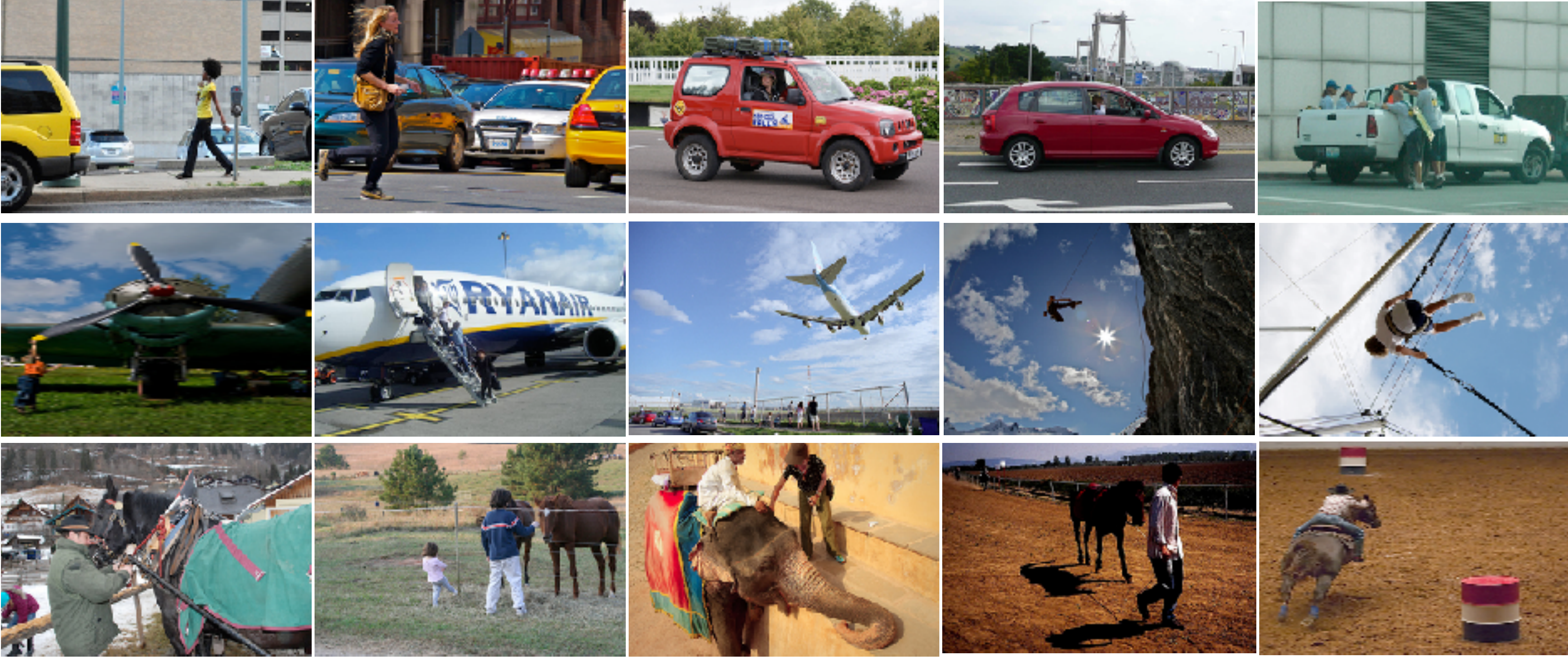}
        \caption{Matching images in the coordinated embedding space with the proposed jWAE.}
        \label{fig:semantic_b}
    \end{subfigure}%
    \caption{\emph{Semantic continuity with Gaussian regularization.} Each row shows images that are  close in the learned embedding space. \emph{(\subref{fig:semantic_a})} Without prior knowledge of matching image pairs, supervised cross-modal methods do not achieve semantic continuity. \emph{(\subref{fig:semantic_b})} Our jWAE based on Gaussian regularization achieves semantic continuity in the coordinated image space in absence of within-domain supervision. 
    }
\label{fig:continuity}
%\vspace{-0.5em}
\end{figure*}

One of the main challenges in multimodal learning is to obtain meaningful latent representations such that they capture semantics that are present across modalities, even if the paired training data does not extensively cover all relevant semantic concepts.
For example, the Flickr30k \cite{YoungLHH14} and COCO \cite{LinMBHPRDZ14} image captioning datasets do not contain matching image pairs.
Lacking within-domain structural constraints, semantic similarity within each modality may not be preserved in the embedding space (\cref{fig:teaser_b,fig:semantic_a}).

We address this problem by learning \emph{semantically continuous} latent representations of images and texts in their respective embedding spaces, \ie multimodal embeddings that encourage a smooth change in the semantics of the input modalities.
We adopt a semi-supervised setting and propose a \emph{joint Wasserstein autoencoder} (jWAE) model, leveraging that regularized autoencoders are known to yield semantically meaningful latent spaces \cite{KingmaW13,wae}.
Specifically, we adopt Gaussian regularization to ensure semantically continuous latent representations of each input modality (\cf \cref{fig:semantic_b}).
In contrast to standard Wasserstein autoencoders (WAEs) \cite{wae}, we share the Gaussian prior across modalities to encourage comparable levels of semantic continuity in both modalities.
Unlike variational autoencoders (VAEs) \cite{KingmaW13}, Wasserstein autoencoders map the input data to a point in the latent space, which allows for the co-ordination of the two modalities through a supervised loss based on matching image-text pairs.
The advantage of the shared Gaussian prior on the two modalities is that their semantic representations are comparable and can be better aligned with supervision as illustrated in \cref{fig:teaser_c}.

We first evaluate our multimodal embeddings of images and texts on cross-modal retrieval and show that they yield state-of-the-art accuracy on the Flickr30k and COCO datasets. 
One of the crucial advantages of the semantically continuous representation from our semi-supervised approach is its generalization capability across datasets. The benefit over the state of the art widens when embeddings of one dataset are evaluated on a related, \emph{previously unseen} dataset. 
Finally, we demonstrate the advantage of our jWAE on phrase localization on the Flickr30k Entities dataset \cite{PlummerWCCHL17}, where we again outperform recent methods from the literature.

\section{Related Work}

\paragraph{Supervised multimodal learning.}
Early work on multimodal learning includes Canonical Correlation Analysis (CCA) \cite{HardoonSS04} and kernel CCA (KCCA) \cite{LaiF00}, which maximize correlation to learn projections of joint embeddings  (\eg, of images and texts). 
However, these methods do not scale to large datasets \cite{MaLF15}.
Deep Canonical Correlation Analysis (DCCA) \cite{AndrewABL13} aims to overcome this scalability issue.
Yet, optimization is challenging as the covariance matrix has to be estimated during training and is prone to over-fitting.

Many recent works formulate embedding multiple domains into a joint space as a learning-to-rank problem \cite{FaghriFKF17,FromeCSBDRM13,GuCJN018,GuoZYLW18,KirosSZ14,wang2018learning,WestonBU11}.
A ranking hinge loss with a margin is used such that matching cross-domain pairs are ranked higher (\ie~are closer in the latent space) than non-matching pairs.
Wang \etal~\cite{wang2018learning} additionally incorporate structural information on the input representations themselves with a domain-specific ranking loss. 
This requires prior knowledge about within-domain matching pairs, which was only available for text. 
Image-text matching has also been studied in a classification setting, employing logistic regression or a softmax with cross-entropy \cite{FukuiPYRDR16,wang2018learning}. 

Gu \etal~\cite{GuCJN018} augment the ranking loss with a conditional generative model framework for cross-modal generation to obtain fine-grained multimodal features.
Harada \etal~\cite{HaradaSMU17} learn an image-text embedding space with a Gaussian prior using generative adversarial networks. 
However, the Gaussian latent prior is applied only on the image modality, and the text distribution is matched to the latent image distribution. 
Moreover, the adversarial framework can suffer from mode collapse. 
Chi \etal~\cite{ChiP18} match image and text embeddings to the label representation.
They assume that image and text have same label, which is limited to tasks where only one concept is required per image.

Wehrmann \etal~\cite{WehrmannB18} improve sentence representations with a character level inception module and \cite{0008WS018,LeeCHHH18} improve image representations for image-text matching models. 
Huang \etal~\cite{0008WS018} use multi-label classification to extract various concepts in images, requiring additional image annotations. 
Lee \etal~\cite{LeeCHHH18} propose an attention mechanism for aligning image regions with words in a sentence. 
This is orthogonal to the underlying multimodal embedding and can be combined with the proposed jWAE framework.

\myparagraph{Semi- and unsupervised multimodal learning.}
Various approaches \cite{NgiamKKNLN11,TsaiHS17} use autoencoders to obtain latent representations. 
Embeddings of the two domains can be aligned using distribution matching constraints \cite{TsaiHS17,WangYXHS17}.
Unlike previous work, which does not encourage any continuity in the latent semantic space across modalities, we use \emph{regularized} autoencoders based on generative models, which enforce the latent embeddings from the encoders to match a prior distribution, thereby yielding continuity in the embedding space.
We use the latent representations obtained from these models as a basis for our multimodal approach.

\begin{table}[t!]
\centering
\scriptsize
\begin{tabularx}{\linewidth}{@{}m{0.25\linewidth}m{0.35\linewidth}m{0.3\linewidth}@{}}
\toprule
\bfseries Text & \bfseries Supervised approach \cite{wang2018learning} & \bfseries Ours (jWAE-MSE) \\
\midrule
Two tan dogs play on the grass near the wall. & \includegraphics[width=0.9\linewidth]{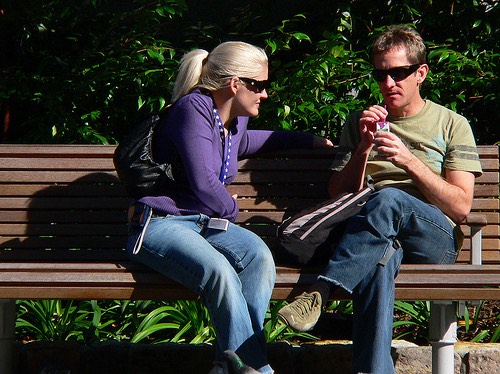} & \includegraphics[width=\linewidth]{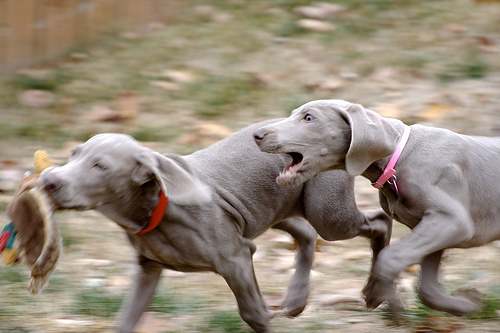}\\
\midrule
 A tennis player wearing white is jumping up to hit a ball. & \includegraphics[width=0.9\linewidth]{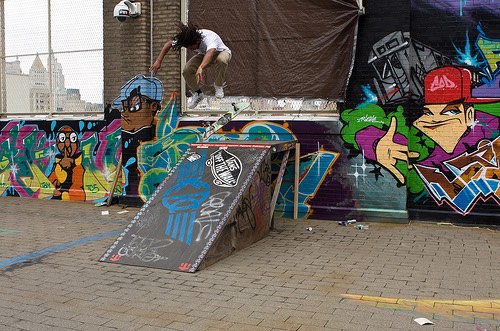} &  \includegraphics[width=0.7\linewidth]{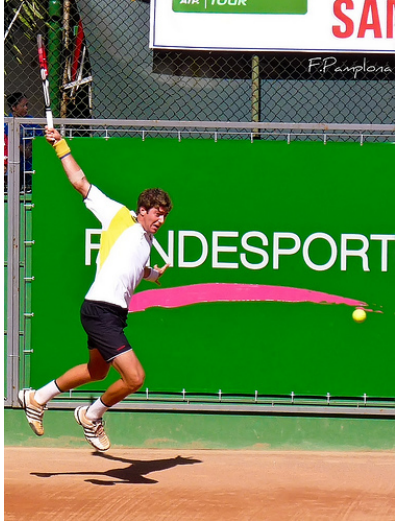} \\
\bottomrule
\end{tabularx}
\caption{Examples of images retrieved for a given sentence on Flickr30k based on embeddings trained on COCO.}\label{table:t2i_cross}
%\vspace{-0.5em}
\end{table}

\myparagraph{Deep generative models}
aim to minimize the difference between the model and the empirical data distribution, and have been successfully applied, \eg, to image generation tasks.
Generative adversarial networks (GANs) \cite{GoodfellowPMXWOCB14} generate the model distribution in a one step procedure where decoders are input with random (Gaussian) noise to construct the data distribution. 
Regularized autoencoders such as variational autoencoders (VAEs)  \cite{KingmaW13} model the data distribution through a two step procedure. 
The empirical distribution is first mapped to a latent space via encoders and then mapped back to the data space via decoders. 
VAEs minimize the reconstruction error between the input and output representation and balance this with the discrepancy between the encoded representation in the latent space and a prior distribution, \eg~a Gaussian, for \emph{each} input. 
Recently, \cite{MakhzaniSJG15,wae} proposed autoencoder-based frameworks where the discrepancy between the encoded distribution of \emph{all} input representations and prior distribution is minimized.
This forces the entire encoded distribution to match the prior.
Such regularization captures the semantics of entire input distribution in a continuous latent representation, desirable for encoding each modality in multimodal learning.

\section{Motivation \& Background}
Current approaches formulate the task of learning multi-modal representations of images and text in an image-text matching framework \cite{FaghriFKF17,WangLL16}, in which a ranking loss is minimized in a fully supervised setting. 
For matching image-text pairs $(x_l,y_l)$ and non-matching images $x_{l'}$ or texts $y_{l'}$ with similarity function $s(x,y)$, the ranking formulation based on the max-margin hinge loss with margin $m$ is defined as 
\begin{align}
\begin{split}
\label{eq:sh}
\mathcal{L}_\text{MH} = \sum_l &\mathcal{\psi}\Big(\max\big[0,m+s(x_l,y_{l'})-s(x_l,y_l)\big]\Big)\\ +&   \mathcal{\psi}\Big(\max\big[0,m+s(x_{l'},y_{l})-s(x_l,y_l)\big]\Big).
\end{split}
\end{align}
Here, $\mathcal{\psi}$ is either the sum of hinge losses over all the negative samples for a given matching pair, or the maximum over all hinge losses. 
The dependence on the choice of negative examples limits the robustness and generalization of the obtained multi-modal embeddings; they fit to the particularities of a dataset, which can be seen from the example retrievals in \cref{table:i2t_cross,table:t2i_cross} on the Flickr30k dataset for an embedding space trained on the COCO dataset. While methods like domain adaptation rely on data from source \emph{and} target datasets to adapt a model to perform better on a specific target dataset \cite{HoffmanTPZISED18,TzengHSD17}, in this work we show the performance of the embeddings where both supervised and unsupervised losses are trained \emph{only} on the source dataset, commonly referred to as \emph{generalization}.

\begin{table}[t!]
\centering
\scriptsize
\begin{tabularx}{\linewidth}{@{}lXX@{}}
\toprule
\bfseries Image &
\bfseries Supervised approach \cite{wang2018learning} &
\bfseries Ours (jWAE-MSE)\\
\midrule
\raisebox{-\totalheight}{\includegraphics[width=0.2\linewidth]{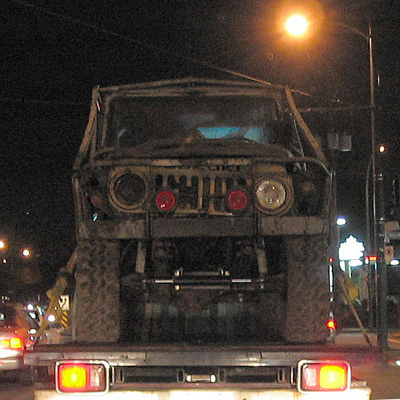}} & \begin{itemize}[nosep,leftmargin=*]\item Chevrolet car on display at a convention. \item Construction in the city at night. \item Two firemen beside their fire engine. \item  An escalator with many people on it, leading out of a tunnel.
\end{itemize} & \begin{itemize}[nosep,leftmargin=*] \item Two firemen beside their fire engine. \item BSR trucks and machinery and workers. \item An old, beat-up jeep being towed away. \item  Construction in the city at night.
\end{itemize}\\[-6pt]
\midrule
\raisebox{-\totalheight}{\includegraphics[width=0.2\linewidth]{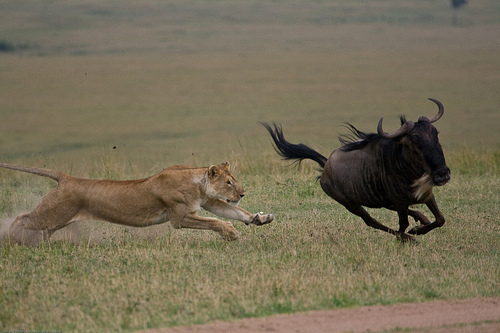}} & \begin{itemize}[nosep,leftmargin=*] \item The dog is running around the cow. \item  A person laying on the ground next to a cow. \item Two children pet horses in a field.  \item  Girl atop horse that is chasing a small longhorn.\end{itemize} & \begin{itemize}[nosep,leftmargin=*] \item A lioness is chasing a black bison across a grassy plain. \item A lioness chases a black animal with horns. \item Two brown dogs play.  \item  Horse jockeys racing on horses in a race.\end{itemize} \\[-6pt]
\bottomrule
\end{tabularx}
\caption{Examples of the top-4 captions retrieved for a given image on Flickr30k based on embeddings trained on COCO.} \label{table:i2t_cross}
%\vspace{-0.5em}
\end{table}

To overcome the limitation of existing methods in expressing semantic coherence within a domain and across multiple domains, we propose to employ Gaussian regularization on the latent distribution.
By virtue of the autoencoder framework, structurally similar input representations are close to each other in the low-dimensional latent space; Gaussian regularization encourages \emph{continuity} in the space of encoded representations. 
Semantic alignment of these spaces is further obtained with supervision. 
We refer to the resultant embeddings as \emph{semantically continuous}.
Our model consists of three main components.
First, each input distribution is mapped to a Gaussian distribution where semantically similar representations are close to each other within the domain.
This is illustrated in \cref{fig:semantic_b}, where images that are close in the embedding space are semantically related.
Second, we share this Gaussian prior across domains, leading to compatible levels of continuity in both domains.
Third, the latent representations of images and text are semantically aligned with supervised information.
As shown in \cref{fig:teaser_c}, images and texts representing similar semantics come closer in the proposed joint embedding space.
Moreover, this offers significantly better generalization across datasets as seen in \cref{table:i2t_cross,table:t2i_cross}, where the example captions retrieved with our semi-supervised approach are semantically more related to the given image than when only employing supervised ranking.

\section{Approach}
We propose a semi-supervised approach for improving the semantic alignment between two modalities, such as images and texts. 
Let $X = \{x_i\}_{i=1}^{N_X}$ and $Y=\{y_j\}_{j=1}^{N_Y}$ denote unpaired input images and texts, respectively. 
Further, let $S = \{\left(x_l, y_l\right)\}_{l=1}^{N_S}$ be matching image and text pairs.
We assume the latent space of each domain to be of dimension $d$.
Encoders $f_v:\mathbb{R}^{d_v} \to \mathbb{R}^d$ and $f_t:\mathbb{R}^{d_t} \to \mathbb{R}^d$ map visual data (images) and text to their respective $d$-dimensional latent spaces. 
$g_v:\mathbb{R}^{d} \to \mathbb{R}^{d_v}$ and $g_t:\mathbb{R}^{d} \to \mathbb{R}^{d_t}$ are the decoders that map the latent representations back to images and text. 
We denote the latent representations of images and texts by $\tilde{v}$ and $\tilde{t}$, respectively.
Next, we let $P_L \sim \mathcal{N}\left(0,\mathit{I}_d\right)$ be a unit Gaussian prior in the $d$-dimensional space with identity covariance matrix $\mathit{I}_d$ and denote the encoded image and text distributions as $F_v= \{f_v\left(x_i\right)\}_{i=1}^{N_X}$ and $F_t=\{f_t\left(y_j\right)\}_{j=1}^{N_Y}$,  respectively. 
Similarly, we define the output (reconstructed) model distributions as $G_v= \{g_v(\tilde{v}_i)\}_{i=1}^{N_X}$ and $G_t= \{g_t(\tilde{t}_j)\}_{j=1}^{N_Y}$. 
Following the common abuse of notation, we let $F_v$ and $F_t$ denote both the encoded activations and the distribution over encodings.

\subsection{Wasserstein autoencoder backbone}
We build the latent representation of each domain, images or text, on Wasserstein autoencoders \cite{wae}. 
We first describe the WAE backbone for the image pipeline and then extend it to text.

Generative models such as VAEs and WAEs minimize the discrepancy between the true data distribution $X$ and the model (reconstructed) distribution $G_v$. 
In such models, latent variables $z$, sampled from a fixed prior distribution $P_L$ in the latent space, are mapped to the space of the original data with parameterized functions $g_v$, and $f$-divergences, such as the KL-divergence or the Jensen-Shannon divergence, between the distributions are minimized.
In high-dimensional spaces, estimating the model distribution directly by sampling in the latent domain would require a large number of samples from the latent distribution.
Therefore, representing the model distribution by random sampling of the latent distribution is computationally expensive.

Variational autoencoders make sampling efficient by introducing a proposal distribution for each $x_i$. 
Specifically, $F_v(x_i)$ is a latent distribution that generates latent representations $z$ likely to produce $x_i$. The VAE minimizes
\begin{align}
\label{eq:vae}
\mathcal{L}_\text{VAE} = \mathcal{D}_\text{KL}\big( F_v(x_i)||P_L\big) - \mathbb{E}_{F_v(x_i)}\big[\log G_v\left(\tilde{v}_i\right)\big],
\end{align}
where the first term encourages the latent variables $z$ over each $x_i$ to match the prior distribution $P_L$. 
While this helps latent samples to be representative of a data point, it does not capture the full underlying true data distribution.
As has been pointed out by \cite{wae} for a Gaussian prior, this results in overlapping Gaussians in the latent space from different input data points. 
This also is the cause of blurry images from VAEs in image generation tasks.
Moreover, mapping the input to a latent \emph{distribution} is problematic when using supervision in the latent space, as intended here.

To model the entire data distribution in the latent space, minimizing $\mathcal{D}(F_v||P_L)$ with $F_v = \int F_v\left(x\right)dX$ is thus desirable \cite{wae}. 
To that end, given the input distribution $X$ and the model distribution $G_v$, Wasserstein autoencoders (WAEs) minimize the optimal transport cost $W_c(X, G_v)$ between the two distributions. Optimal transport with a cost function $c\left(s,t\right): \mathbb{R}^d \times \mathbb{R}^d \to \mathbb{R}_+$ is defined as 
\begin{align}
	W_c(X,G_v) = \inf_{\gamma \in \Gamma(X,G_v)}\mathbb{E}_{\left(s,t\right) \sim \gamma} \big[c(s,t)\big],
	\label{eq:wdist}
\end{align}
where $\Gamma(X,G_v)$ is the set of all possible joint distributions (couplings) of $(s,t)$ whose marginals are $X$ and $G_v$, respectively. 
In generative models with a deterministic mapping from the latent distribution $P_L$ to $G_v$, the optimal transport cost between $X$ and $G_v$ reduces to finding a conditional distribution $F_v$ such that $\int F_v(x)dX = P_L$ \cite{bousquet2017optimal,wae}. 
This constraint is enforced as a regularization term by minimizing $\mathcal{D}\left( F_v||P_L \right)$. 
This yields Wasserstein autoencoders as
\begin{align}
\mathcal{L}_\text{WAE} = \inf_{F_v \in \mathcal{F}} \mathbb{E}_{X}\mathbb{E}_{F_v}\big[c(x,g_v(x))\big] + \lambda\mathcal{D}\big( F_v||P_L \big).
\label{eq:wae}
\end{align}
Here, $c\left(s,t\right):  \mathbb{R}^d \times \mathbb{R}^d \to \mathbb{R}_+$ is a cost function and $\mathcal{F}$ is a set of probabilistic encoders without any constraints (non-parametric), which model $p(z|x)$.
We take it to be the set of all fully-connected networks with fixed size. 

\begin{figure*}[t!]
 \centering
 %\begin{minipage}[t]{0.18\textwidth}
   % \vspace{0pt}
  %\end{minipage}
 %\hfill
 %\begin{minipage}[t]{0.75\textwidth}
    %\vspace{0pt}
    \includegraphics[width=0.75\linewidth]{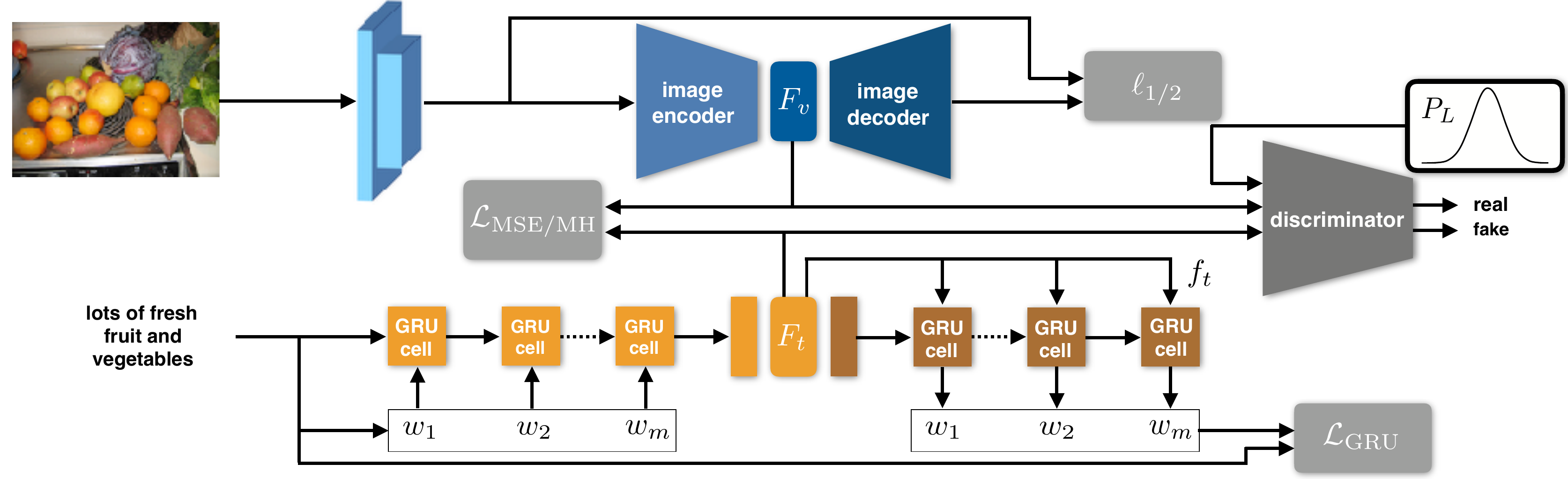}
 %\end{minipage}
  \caption{Architecture of our joint Wasserstein Autoencoder.}
   \label{fig:archi}
\end{figure*}

Choosing $\mathcal{D}$ as the  KL-divergence between $F_v$ and $P_L$ would require closed forms of $F_v$ and $P_L$.
Since the closed form of $F_v$ is not available, we instead minimize the Jensen-Shannon divergence between the prior and encoded latent distributions $\mathcal{D}_\text{JS}\left(F_v||P_L\right)$ using a GAN-based formulation, which allows to conveniently use samples from the distribution. 
We refer to $\mathcal{D}_\text{JS}$ as Gaussian regularization since we consider the prior to be a unit Gaussian.

Note that the popular Wasserstein GANs \cite{ArjovskyCB17} minimize the optimal transport cost (Eq.~\ref{eq:wdist}) between the data distribution and the model distribution using the dual formulation of the optimal transport cost directly from the latent space to the data space.
That is, they do not have an encoder that maps the input to latent representations.
However, for learning coordinated representations of two input data distributions with a shared prior, an encoder that maps data points to the latent space is desirable.
We thus build on WAEs here.  

\subsection{Joint WAE}
For learning coordinated representations, we are now interested in formulating continuous $d$-dimensional embedding spaces for each modality, which are aligned through constraints.
To this end, we propose to share the prior on the latent representations.
Specifically, the latent representations of each domain are constrained to be close to the same unit Gaussian distribution;
that is we minimize the discrepancy $ \mathcal{D}\left(F_v||P_L\right)$ and $\mathcal{D}\left(F_t||P_L\right)$ between the encoded representations of each modality and the Gaussian prior $P_L$. 

We now formulate our joint Wasserstein autoencoder in terms of a classical encoder-decoder setting.
To that end, we first note that when $f_v$ and $g_v$ in a regular WAE (Eq.~\ref{eq:wae}) are parameterized with encoders and decoders in a deep neural network framework, the first term of \cref{eq:wae} reduces to minimizing the reconstruction error between the true input representation and the decoded representation.

To apply the WAE formulation to sentences, we first extend it to gated recurrent units (GRUs) \cite{SchusterP97} where $f_t$ is the encoded output of the GRU encoder and Gaussian regularization is applied to the encoded distribution, $F_t$, of all the sentences. 
$f_t$ serves as the input hidden state to the GRU decoder and the reconstruction loss for the sentence is 
\begin{align}
\label{eq:ours-gru}
\mathcal{L}_\text{GRU} = -\sum_{m=0}^{M-1} \log p_{g_t}\Big(w^j_m\Big|w^j_{0:m-1},f_t(y_j);g_t\Big),
\end{align}
where $w^j_m$ is the ground truth word and $p_{g_t}(w^j_m|w^j_{0:m-1},f_t)$ is the output probability of word $w^j_m$ in sentence $y_j$ given the decoder $g_t$ and hidden state $f_t$.

Recalling that $F_v$ and $F_t$ denote the encoded distribution of images and text, respectively, we formulate the unsupervised part of our joint WAE loss as
\begin{align}
\label{eq:ours}
\begin{split}
\mathcal{L}_\text{jWAE} =\:& \lambda_1\sum_{i=1}^{N_X}\Big\|x_i-g_v\big(f_v(x_i)\big)\Big\|  \\
 -\:& \lambda_2\sum_{j=1}^{N_Y}\sum_{m=0}^{M-1} \log p_{g_t}\left(w^j_m\Big|w^j_{0:m-1},f_t(y_j);g_t\right) \\ +\:& \lambda_3 \mathcal{D}_\text{JS}(F_v||P_L) + \lambda_4 \mathcal{D}_\text{JS}(F_t||P_L).\raisetag{12pt}
\end{split}
\end{align}
Here, $\{\lambda_1, \ldots, \lambda_4\}$ are the regularization parameters and $\|.\|$ is $\ell_1$ or $\ell_2$ norm. 
Modality specific reconstruction error terms are crucial components in preventing mode collapse and encouraging diversity in the latent representations of each domain.
For sentences encoded with fully connected encoders using pre-trained sentence encodings like average of word2vec \cite{word2vec}, we use an analogous formulation with the mean-squared error as the reconstruction loss. 

Gaussian regularization itself does not induce any semantic coupling between the cross-modal distributions.
To ensure that the latent spaces not only have a compatible continuity but rather align semantic concepts across modalities, we add a supervised loss minimizing the distance between latent representations of matching image-text pairs.
Embeddings of the two domains can now be directly aligned with the mean-squared error
\begin{equation}
\mathcal{L}_\text{MSE} = \frac{1}{N_S}\sum_{l=1}^{N_S}\big\|f_v(x_l)-f_t(y_l)\big\|^2.
\end{equation}
The overall loss function of our approach is then given as
\begin{equation}
\label{eq:jWAE-MSE}
\mathcal{L}_\text{jWAE-MSE} = \mathcal{L}_\text{jWAE} + \mathcal{L}_\text{MSE}.
\end{equation}
Alternatively, we also show the effect of Gaussian regularization with the max-margin hinge loss \cite{FaghriFKF17} from \cref{eq:sh} as supervised loss function 
\begin{equation}
\label{eq:jWAE-MH}
\mathcal{L}_\text{jWAE-MH} = \mathcal{L}_\text{jWAE} + \mathcal{L}_\text{MH}. 
\end{equation}
The overall model architecture is illustrated in \cref{fig:archi}.

\myparagraph{Implementation.}
The outputs from the encoders (\ie the encoded distributions $F_v$ and $F_t$) along with $z \sim P_L$ are input to the discriminator.
The number of samples from the prior distribution $P_L$ equals the sum of the samples output by the two encoders. 
The discriminator distinguishes between the encoded distributions and the joint Gaussian prior. 
Note that we implement a single discriminator network for two generator networks, which makes our architecture computationally efficient. 
This is possible as the same prior distribution is used for images and texts.
The discriminator is a fully-connected three layer neural network with leaky ReLU non-linearities after the first two layers, which enables a better flow of gradients during optimization \cite{SalimansGZCRCC16}.
The generator network (the encoder) of each pipeline tries to ``fool" the discriminator network by generating encodings close to the Gaussian prior distribution.

In general, both encoders and decoders consist of two fully connected layers; ReLU non-linearities are applied after the first layers. 
For the text pipeline based on GRUs, the encoder is a bi-directional GRU with two layers. 
The output of the GRU is encoded in the latent space after application of a linear fully-connected layer. 
The decoder is a uni-directional GRU with word dropout encouraging meaningful latent representations of sentences.

\begin{table*}[t!]
    \centering
    \footnotesize
    \begin{tabularx}{\textwidth}{@{}lXS[table-format=2.1]S[table-format=2.1]S[table-format=2.1]S[table-format=2.1]S[table-format=2.1]S[table-format=2.1]S[table-format=2.1]S[table-format=2.1]S[table-format=2.1]S[table-format=2.1]S[table-format=2.1]S[table-format=2.1]@{}}
    \toprule
    \multicolumn{2}{@{}c}{}& \multicolumn{6}{c}{Recall on Flickr30k} & \multicolumn{6}{c}{Recall on COCO} \\
    \cmidrule(lr){3-8}\cmidrule(lr){9-14}
    \multicolumn{2}{@{}c}{}& \multicolumn{3}{c}{Image-to-text} &\multicolumn{3}{c}{Text-to-image} & \multicolumn{3}{c}{Image-to-text} &\multicolumn{3}{c}{Text-to-image}\\
    \cmidrule(lr){3-5}\cmidrule(l){6-8}\cmidrule(lr){9-11}\cmidrule(l){12-14}
    & {Method (Features)} & {@1} & {@5} & {@10} & {@1} & {@5} & {@10} & {@1} & {@5} & {@10} & {@1} & {@5} & {@10} \\
    \midrule
    \multirow{4}{1.25cm}{Without\\ Negative\\ Sampling}
                 & {CCA (VGG+HGLMM) \cite{KleinLSW15}}                      & 34.4 & 61.0 & 72.3 & 24.4 & 52.1 & 65.6 & 37.7 & 66.6 & 79.1 & 24.9 & 58.8 & 76.5 \\
                 & {CCA (Mean Vector) \cite{KleinLSW15}}                    & 24.8 & 52.5 & 64.3 & 20.5 & 46.3 & 59.3 & 33.2 & 61.8 & 75.1 & 24.2 & 56.4 & 72.4 \\
                 & {Sim.~Network (VGG+HGLMM) \cite{wang2018learning}}             & 16.6 & 38.8 & 51.0 &  7.4 & 23.5 & 33.3 & 30.9 & 61.1 & 76.2 & 14.0 & 30.0 & 37.8 \\
                 & {MMD-MSE (VGG+w2v) \cite{TsaiHS17}}                                      & 35.3 & 60.8 & 72.8 & 16.5 & 40.2 & 52.9 & 42.9 & 73.8 & 84.4 & 19.6 & 50.2 & 66.6 \\
    \midrule
    \multirow{8}{1.25cm}{With\\ Negative\\ Sampling}
                 & {Emb.~Network (VGG+w2v) \cite{wang2018learning}}    & 35.7 & 62.9 & 74.4 & 25.1 & 53.9 & 66.1 & 40.7 & 74.2 & 85.3 & 33.5 & 68.7 & 83.2 \\
                 & {Emb.~Network (VGG+HGLMM) \cite{wang2018learning}}  & 40.3 & 68.9 & 79.9 & 29.7 & 60.1 & 72.1 & 50.1 & 79.7 & 89.2 & 39.6 & 75.2 & 86.9 \\
                  & {2WayNet  (VGG+HGLMM) \cite{EisenschtatW17}} & 49.8 & 67.5 & {--} & 36.0 & 55.6 & {--} & 55.8 & 75.2 & {--} & 39.7 & 63.3 & {--} \\
                 & {VSE++ (Resnet+GRU) \cite{FaghriFKF17}}                   & 52.9 & 80.5 & 87.2 & 39.6 & 70.1 & 79.5 & 64.6 & 90.0 &  95.7 & 52.0   & 84.3 & 92.0 \\
                  & {GXN (ResNet+GRU) \cite{GuoZYLW18}}                  & 56.8 &{--}&  89.6 & 41.5   & {--}& 80.1 & 68.5 & {--} &  97.9 & 56.6   & {--} & 94.5 \\
                 & {SCO (Resnet+GRU) \cite{0008WS018}}                  & 55.5& 82.0 & 89.3 & 41.1 & 60.5 & 80.1 & 69.9& 92.9 & 97.5 & 56.7 & \bfseries 87.5 & \bfseries 94.8 \\
                   & {SCAN t-i (ResNet+GRU) \cite{LeeCHHH18}}                                & 61.8 & 87.5 & 93.7 & 45.8 & 74.4& 83.0 &  70.9 & \bfseries 94.5 & 97.8 & 56.4 & 87.0 & 93.9 \\
                   & {SCAN i-t (ResNet+GRU) \cite{LeeCHHH18}}                                & 67.9 & 89.0 & 94.4 & 43.9 & 74.2& 82.8 & 69.2 & 93.2 & 97.5 & 54.4 & 86.0 & 93.6\\

    \midrule
                 & {MSE (VGG+w2v)}                                   & 33.9 & 61.3 & 73.2 & 15.9 & 40.3 & 52.8 & 45.0 & 76.5 & 87.0 & 21.3 & 52.3 & 68.1 \\
                 & {jWAE-MSE (VGG+w2v)}                                         & 35.7 & 61.6 & 73.6 & 17.3 & 41.8 & 55.3 & 43.2 & 75.1 & 85.7 & 21.5 & 53.5 & 69.3 \\
                  & {jWAE-MH (VGG+w2v)}                                        & 35.4 & 62.5 & 74.4 & 24.1 & 50.4 & 62.6 & 44.2 & 77.4 & 87.7 & 31.3 & 66.0 & 81.3 \\
                 & {jWAE-MSE (VGG+HGLMM)}                                       & 40.3 & 66.3 & 77.2 & 20.3 & 46.5 & 58.9 & 50.3 & 79.4 & 88.3 & 25.2 & 57.5 & 73.3 \\
                 & {jWAE-MH (ResNet+GRU)}                                   & {53.9} & {82.2} & {87.4} & {40.7} & {72.4} & {81.9} & 66.6    & 91.4 &  96.6 &  53.1   & 84.5 & 92.0 \\
                 & {jWAE-MH+SCAN t-i (ResNet+GRU)}                                   &  64.0 &  89.4  &  \bfseries 95.2 & \bfseries  47.1 &  \bfseries 75.9 & \bfseries  84.0 & \bfseries 72.0 & \bfseries94.5  & \bfseries 98.3 & \bfseries 57.1 & \bfseries 87.5 & 94.1 \\
                 & {jWAE-MH+SCAN i-t (ResNet+GRU)}                                   & \bfseries 68.5 & \bfseries 89.6  & 94.2 & 44.2 & 74.2 & 83.2 & {69.8} & {93.3} & {98.1} & {54.6} & {86.1} & {93.2} \\
    \bottomrule
    \end{tabularx}
 \vspace{-0.1 cm}
 \caption{{Cross-modal retrieval results (in \%) on the Flickr30k dataset \cite{YoungLHH14} as well as the COCO dataset \cite{LinMBHPRDZ14} with 1000 test images.}}
 \label{table:flickr-coco}
 \vspace{-0.5em}
 \end{table*}

\section{Experiments}
To show the applicability of our jWAE framework across \emph{different tasks}, we evaluate the learned multimodal embeddings on cross-modal retrieval and phrase localization.

\subsection{Cross-modal retrieval}

\paragraph{Visual input.} 
We consider pretrained VGG-19 \cite{SimonyanZ14a} and  ResNet-152 \cite{HeZRS16} models for image features. 
For VGG-19, we extract the 4096-dimensional feature vector from the first fully connected layer and for ResNet-152 the 2048-dimensional activations from the fully connected layer.

\myparagraph{Textual input.}
Following \cite{wang2018learning}, we use the mean of 300-dimensional word2vec \cite{word2vec} features of the words in the sentence.
Alternatively, we use nonlinear Fisher vectors from a hybrid Gaussian-Laplacian mixture model (HGLMM)
\cite{KleinLSW15}.
For GRU, one-hot encodings of the words are projected with an embedding layer, which is initialized either randomly or with pre-trained word2vec embeddings.

\myparagraph{Datasets.} 
Two popular benchmark datasets for evaluating multimodal visual and textual representations are Flickr30k and COCO \cite{FaghriFKF17,KleinLSW15,wang2018learning}.
Flickr30k \cite{YoungLHH14} is comprised of 31783 images with five captions per image. 
We use the splits of \cite{KarpathyF17,WangLL16}; 
validation and test splits consist of 1000 images with 5 captions each. 
The remaining images are used for training. 
COCO \cite{LinMBHPRDZ14} is larger and more diverse than Flickr30k. 
It consists of 82783 training images with five captions each.
5000 images from the validation set are retained for validation purposes and the remaining 30504 images are used for training.
Similar to \cite{KarpathyF17,KleinLSW15,WangLL16}, we use 1000 images with their captions in the test split. 

\myparagraph{Network training.} 
We train the network (see Appendix B for architectural details) using the Adam optimizer with a learning rate of 1E-4. 
The discriminator is trained with a learning rate of 5E-5.
The batch size is taken as 64 or 128.
For jWAE-MSE, the regularization parameters are set to $\lambda_1=\lambda_2=1.0$ for the reconstruction terms in \cref{eq:ours} and $\lambda_3=\lambda_4=0.2$ for the Gaussian regularization. For jWAE-MH, the parameters are $\lambda_1=0.5$, $\lambda_2=0.005$,  $\lambda_3=\lambda_4=0.01$ for most of the experiments.  

\myparagraph{Evaluation metric.} 
In cross-modal retrieval tasks, Recall@$K$ is a standard performance measure and defined as the fraction of instances for which their ground truth is in the top-$K$ based on a similarity score (cosine similarity).

\myparagraph{Baselines \& methods.}
We compare the accuracy of our approach against several state-of-the-art methods for image-to-text and text-to-image retrieval, particularly the Embedding Network of \cite{wang2018learning,WangLL16} and VSE++ \cite{FaghriFKF17}, which use different formulations of the ranking loss, and the attention-based Stacked Cross Attention Network (SCAN) \cite{LeeCHHH18}. 
For evaluation with SCAN, we integrate the jWAE-MH framework with the best-performing setting on the respective dataset.
To compare our approach against methods without negative sampling, we also include the Similarity Network of \cite{wang2018learning} and the Canonical Correlation Analysis (CCA) approach of \cite{KleinLSW15}.
For our jWAE framework, we demonstrate the effect of Gaussian regularization with supervision through the mean-squared error (jWAE-MSE,  Eq.~\ref{eq:jWAE-MSE}) as well as a margin-based hinge loss (jWAE-MH, Eq.~\ref{eq:jWAE-MH}). 
We extend jWAE for the SCAN method (jWAE-MH+SCAN \mbox{t-i}/\mbox{i-t}) where in i-t attention is applied on words with respect to each image region and in t-i image regions are attended with respect to each word in the sentence.
We also include results from a MMD loss for matching text and image distributions (MMD-MSE)  \cite{TsaiHS17} and an ablation of our method without the Gaussian regularization (MSE), both with the usual reconstruction error for autoencoders.

 \begin{table}[t!]
  \centering
  \footnotesize
  \begin{tabularx}{\columnwidth}{@{}XS[table-format=2.1]S[table-format=2.1]S[table-format=2.1]S[table-format=2.1]@{}}
    \toprule
    & \multicolumn{2}{c}{Image-to-text} &\multicolumn{2}{c@{}}{Text-to-image}\\
    \cmidrule(lr){2-3}\cmidrule(l){4-5}
    Method (all ResNet+GRU) & \multicolumn{1}{c}{R@1} &\multicolumn{1}{c}{R@5}&\multicolumn{1}{c}{R@1} &\multicolumn{1}{c@{}}{R@5} \\
    \midrule
    VSE++ \cite{FaghriFKF17} (baseline) & 64.6 & 90.0  & 52.0   & 84.3\\
    VAE-MH  &57.8 &86.9 &46.3 & 80.3 \\
    jWAE-MH  & \bfseries 66.6 & \bfseries 91.4 & \bfseries 53.1 & \bfseries 84.5   \\
    \bottomrule
  \end{tabularx}
  \caption{Comparison of jWAE with VAE on the COCO dataset.
  }
  \label{table:vae}
 \end{table}

 \begin{table}[t!]
    \centering
    \footnotesize
    \begin{tabularx}{\columnwidth}{@{}XS[table-format=2.1]S[table-format=2.1]S[table-format=2.1]S[table-format=2.1]@{}}
    \multicolumn{5}{@{}c@{}}{ \small Flickr30k(Train) $\Rightarrow$ \small COCO(Test) }\\
    \toprule
    & \multicolumn{2}{c}{Image-to-text} &\multicolumn{2}{c@{}}{Text-to-image}\\
    \cmidrule(lr){2-3}\cmidrule(l){4-5}
    Method& {R@5} & {R@10} & {R@5} & {R@10} \\
    \midrule
    CCA (VGG+w2v) \cite{KleinLSW15}  & 13.3 & 20.1 & 10.3 & 16.0 \\
    Embed.~Network (VGG+w2v) \cite{wang2018learning}  & 37.6 & 49.5 & 32.5 & 45.8 \\
    MSE (VGG+w2v)  & 40.1 & 52.9& 21.7 & 33.8\\
    SCAN (ResNet+GRU)~\cite{LeeCHHH18} & 59.2 & 70.0 & 53.2 & 66.6\\\midrule
    jWAE-MSE (VGG+w2v)  & 42.8 & 55.3 & 28.3 & 40.9 \\
    jWAE-MH (VGG+w2v)  & 43.4 &  58.4 & 34.5 &   49.2 \\
    jWAE-MH+SCAN (ResNet+GRU)& \bfseries 64.4 & \bfseries76.9 & \bfseries55.2 & \bfseries 68.7\\
    \bottomrule
    \end{tabularx}
    \caption{Generalization results of models trained only on the Flickr30k training set and evaluated on the COCO test set.
    }
    \label{table:flickr2coco}
    \vspace{-0.5em}
 \end{table}

\myparagraph{Results.}  In \cref{table:flickr-coco}, we show the results of our method for cross-modal retrieval on the Flickr30k and COCO datasets. 

Our jWAE framework leads to competitive results compared to the current state of the art in image-to-text retrieval. For example, jWAE-MH outperforms VSE++ with respect to top-1 recall by $1.0\%$ points on Flickr30k and by $2.0\%$ points on COCO. For the Embedding Network with VGG+w2v features, jWAE-MH has $3.5\%$ better accuracy on COCO. 
Our semi-supervised representations also improve the top-1 recall of SCAN t-i by $2.2\%$ and $2.1\%$ points on Flickr30k and COCO, respectively.
For text-to-image retrieval, improving the top-1 recall is more challenging.
Yet, we also achieve an improvement in top-1 recall of $1.3\%$ and $0.7\%$ points over SCAN t-i, and an improvement of $1.1\%$ for VSE++ on Flickr30k and COCO.
This shows that irrespective of the network architecture and complexity of input features, our jWAE improves the current state of the art. 
To the best of our knowledge, our jWAE framework improves (over) the currently leading cross-modal retrieval methods.

We observe that the recall for text-to-image retrieval of jWAE-MSE is not as competitive and is comparable to methods that do not use negative sampling. 
This can be attributed to the nature of the datasets where five sentences compete for the same image. Moreover,  given a sentence there can be many images that can be described reasonably well by the sentence.
jWAE-MH bridges the gap between maximizing top-K recall by ranking matching image-text pairs higher than non-matching pairs \emph{and} improving accuracy of the embeddings by encouraging semantic continuity.

In \cref{table:vae}, we additionally compare jWAE-based Gaussian regularization to traditional VAEs using VSE++ as baseline. 
While jWAE-MH improves the accuracy of the VSE++ baseline, VAE-MH results in a decrease of top-K recall.
The reason is that VAEs map \emph{each} input to a Gaussian latent \emph{distribution}. 
This hinders the application of a point-wise supervised loss in the latent space. 
In order to match the latent representations of two modalities with the MH loss, we instead require mapping to a \emph{point} in the latent space. 
jWAE enforces global Gaussian priors on the latent distributions while mapping the input to a point in the embedding space. 
Therefore and unlike VAEs, jWAEs are suitable for semi-supervised learning of joint embeddings.

\begin{table}[t!]
  \centering
  \footnotesize
  \begin{tabularx}{\columnwidth}{@{}XS[table-format=2.1]S[table-format=2.1]S[table-format=2.1]S[table-format=2.1]@{}}
    \multicolumn{5}{@{}c@{}}{\small COCO(Train) $\Rightarrow$ \small Flickr30k(Test) }\\
    \toprule
    & \multicolumn{2}{c}{Image-to-text} &\multicolumn{2}{c@{}}{Text-to-image}\\
    \cmidrule(lr){2-3}\cmidrule(l){4-5}
    Method& \multicolumn{1}{c}{R@5} &\multicolumn{1}{c}{R@10}&\multicolumn{1}{c}{R@5} &\multicolumn{1}{c@{}}{R@10} \\
    \midrule
    Embed.~ Network (VGG+w2v) \cite{wang2018learning} & 33.7& 45.5 & 8.4 & 12.2  \\
    MSE  (VGG+w2v) &43.6 & 56.9 & 24.3 & 34.2\\
    SCAN (ResNet+GRU)~\cite{LeeCHHH18} & 73.7&82.6 & 61.3 & 72.4\\\midrule
    jWAE-MSE  (VGG+w2v) &  48.5 & 60.0 &  29.1 &  40.7 \\
    jWAE-MH  (VGG+w2v) &  51.1 &    63.3  &   37.0 &  49.1  \\
     jWAE-MH+SCAN (ResNet+GRU) & \bfseries 80.0& \bfseries 87.0 & \bfseries 66.7 & \bfseries 75.9\\
    \bottomrule
  \end{tabularx}
  \caption{Generalization results of model trained only on the COCO training set and evaluated on the Flickr30k test set.
  }
  \label{table:coco2flickr}
  \vspace{-0.5em}
\end{table}
 
To show that our method learns meaningful representations with continuity in the latent space, we test the cross-dataset generalization capability of our method against various retrieval approaches: 
CCA \cite{KleinLSW15}, the Embedding Network \cite{wang2018learning}, and SCAN \cite{LeeCHHH18}.
For cross-dataset generalization, the model is trained on the training set of one dataset, \eg~COCO (Flickr30k), and tested on the test set of another dataset, \ie~Flickr30k (COCO). Note, we do not train to improve the accuracy on specific target dataset.
We find that previous methods based on global representations \cite{wang2018learning} have low generalization performance with top-10 recall as low as 12.2\% when testing on Flickr30k for a model trained on COCO. Fine-grained representations based on attention \cite{LeeCHHH18} generalize better compared to \cite{wang2018learning}.
Following \cref{table:flickr2coco,table:coco2flickr}, the jWAE-MH framework significantly improves the generalization across datasets further, owing to the semantic continuity from the Gaussian regularization.
For image-to-text and text-to-image retrieval we improve the top-5 recall by $5.2\%$ and $2.0\%$ points, respectively, generalizing from Flickr30k to COCO and by $6.3\%$ and $5.4\%$ points, respectively, for generalizing from COCO to Flickr30k.  

In \cref{fig:continuity}, we show modality-specific semantic continuity, where, \eg, in the third row, for an image with ``a man and a horse'', matching images retrieved by our approach are of `person-animal interaction' whereas for the supervised approach \cite{wang2018learning} matching images show the concept `person'. 
Similarly, for cross-modal semantic continuity in \cref{table:t2i_cross}, \cite{wang2018learning} retrieves an image with a concept `two' while jWAE is able to retrieve the image representative of the given sentence ``Two tan dogs play on grass near the wall". We provide additional qualitative examples in Appendix C.

In the Appendix A, we additionally study the accuracy under limited supervision.
%There, we also find favorable generalization properties.

\subsection{Phrase localization}
We next analyze the benefit of our jWAE framework for phrase localization on the Flickr30k Entities dataset \cite{PlummerWCCHL17}.
Phrase localization associates (grounds) a phrase to a region in the image using bounding boxes \cite{ChenKGN17,RohrbachRHDS16,wang2018learning,YehXHDS17}. 
Following \cite{wang2018learning}, we formulate phrase localization as a retrieval problem where given an image and a phrase from its associated sentence, the phrase is mapped to the regions in the image.
Bounding box proposal regions are extracted with Edge Box \cite{zitnick2014edge}. 
Since we are mainly interested in evaluating the quality of our multimodal embeddings rather than the specific task, we compared to other embedding-based approaches \cite{RohrbachRHDS16,wang2018learning}.
Additionally, we  integrate our jWAE framework with Conditional Image Text Embedding (CITE) \cite{Plummer18}, which builds on top of the embeddings from the Similarity Network. 
We also include \cite{PlummerMCHL17}, which uses additional image and language constraints, and \cite{YehXHDS17}, which considers all possible bounding boxes based on image concepts like segmentation, word priors, and detection scores.

\myparagraph{Dataset and input.}
The Flickr30k Entities dataset \cite{PlummerWCCHL17} augments the captions of images in Flickr30k with 244k mentions of distinct entities across sentences. The mentions are associated with 276k bounding boxes.
%Experiements and Results
Similar to \cite{Plummer18,RohrbachRHDS16,wang2018learning}, we extract 4096-dimensional visual features from  Fast R-CNN \cite{Girshick15}, finetuned on the PASCAL VOC 2007--2012 datasets \cite{EveringhamGWWZ10}.
We use proposal regions with IoU $\geq$ 0.7 as a positive region for a phrase during training.
For encoding phrases, PCA is applied to HGLMM features to reduce the dimensionality to 6000 \cite{KleinLSW15}.

\myparagraph{Results.} 
We compare our method with \cite{FukuiPYRDR16,RohrbachRHDS16,wang2018learning} where multimodal embeddings are evaluated for phrase localization. 
Following these methods, we use 200 or 500 Edge Box proposals per image. An IoU of at least 0.5 is required for a proposal region to match the ground truth bounding box for a phrase.
As shown in Table \ref{table:phraseloc}, our method outperforms previous multimodal embedding networks for the phrase localization task by 1.5\% for top-1 recall. 
The gap compared to \cite{wang2018learning} widens to around 5\% for the top-5 and top-10 recall.
Moreover, using jWAE as the embedding framework in CITE \cite{Plummer18} similarly improves the top-1 recall by 1.2\% with new state of the art results for phrase localization.
This again highlights the improved accuracy of the embeddings obtained from our semi-supervised jWAE approach. Please see Appendix D for additional qualitative results.

\begin{table}[t!]
    \centering
    \footnotesize
    \begin{tabularx}{\columnwidth}{@{}XS[table-format=2.1]S[table-format=2.1]S[table-format=2.1]@{}}
        \toprule
        Methods & {R@1} & {R@5} & {R@10} \\
        \midrule
        MCB \cite{FukuiPYRDR16}                    & 48.7 & {--} & {--} \\
        GroundeR \cite{RohrbachRHDS16}             & 47.8 & {--} & {--} \\
        Embedding Network \cite{wang2018learning}  & 51.0  & 70.4 & 75.5 \\
        Similarity Network \cite{wang2018learning} & 51.0  & 70.3 & 75.0 \\
        SPC \cite{PlummerMCHL17} & 55.4& {--} & {--} \\
        IGOP \cite{YehXHDS17} & 53.9 & {--} & {--} \\
        CITE \cite{Plummer18} & 59.2 & {--} & {--} \\
        \midrule
        jWAE-MSE                                       &  52.5 & \bfseries 75.0 & \bfseries 81.5 \\
        jWAE-MSE+CITE                                & \bfseries 60.4 & {--} & {--}\\
        \bottomrule
    \end{tabularx}
    \caption{{Phrase localization on the Flickr30k Entities dataset \cite{PlummerWCCHL17}.}}
    \label{table:phraseloc}
 \end{table}
\section{Conclusion}
We presented a novel joint Wasserstein autoencoder framework for modeling continuous multimodal representations of images and texts with Gaussian regularization, allowing to better capture the semantic structure in latent representations.
Our experiments show that our multimodal embeddings push the current state of the art under full supervision.
A key advantage of our method is its generalization capability across datasets, where it significantly outperforms recent methods.
We thus believe our semi-supervised approach provides an important step toward learning generalizable multimodal representations, which are a crucial component, \eg, for image captioning \cite{FromeCSBDRM13} in the real world.

\clearpage
{\small 
\myparagraph{Acknowledgement.} This work has been supported by the German Research Foundation as part of the Research Training Group \emph{Adaptive Preparation of Information from Heterogeneous Sources (AIPHES)} under grant No.~GRK 1994/1.}

{\small
\bibliographystyle{ieee_fullname}
\bibliography{shortbib}
}

\clearpage
\appendix

\section*{Appendix A. jWAE with Limited Supervision}

\begin{figure}[t!]
\begin{subfigure}{.49\linewidth}
  \includegraphics[width=\textwidth]{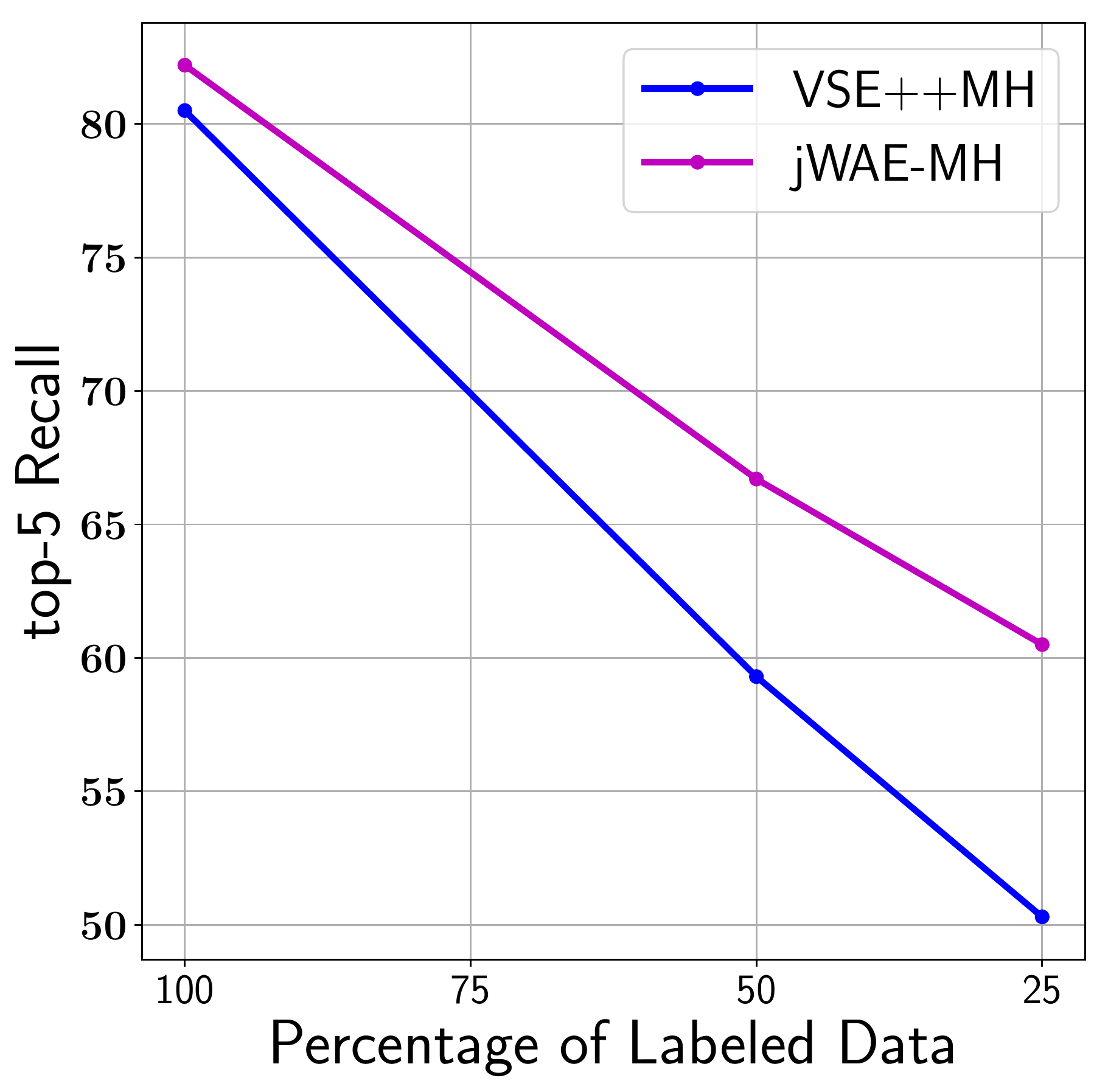}
\end{subfigure}%
\begin{subfigure}{.49\linewidth}
  \includegraphics[width=\textwidth]{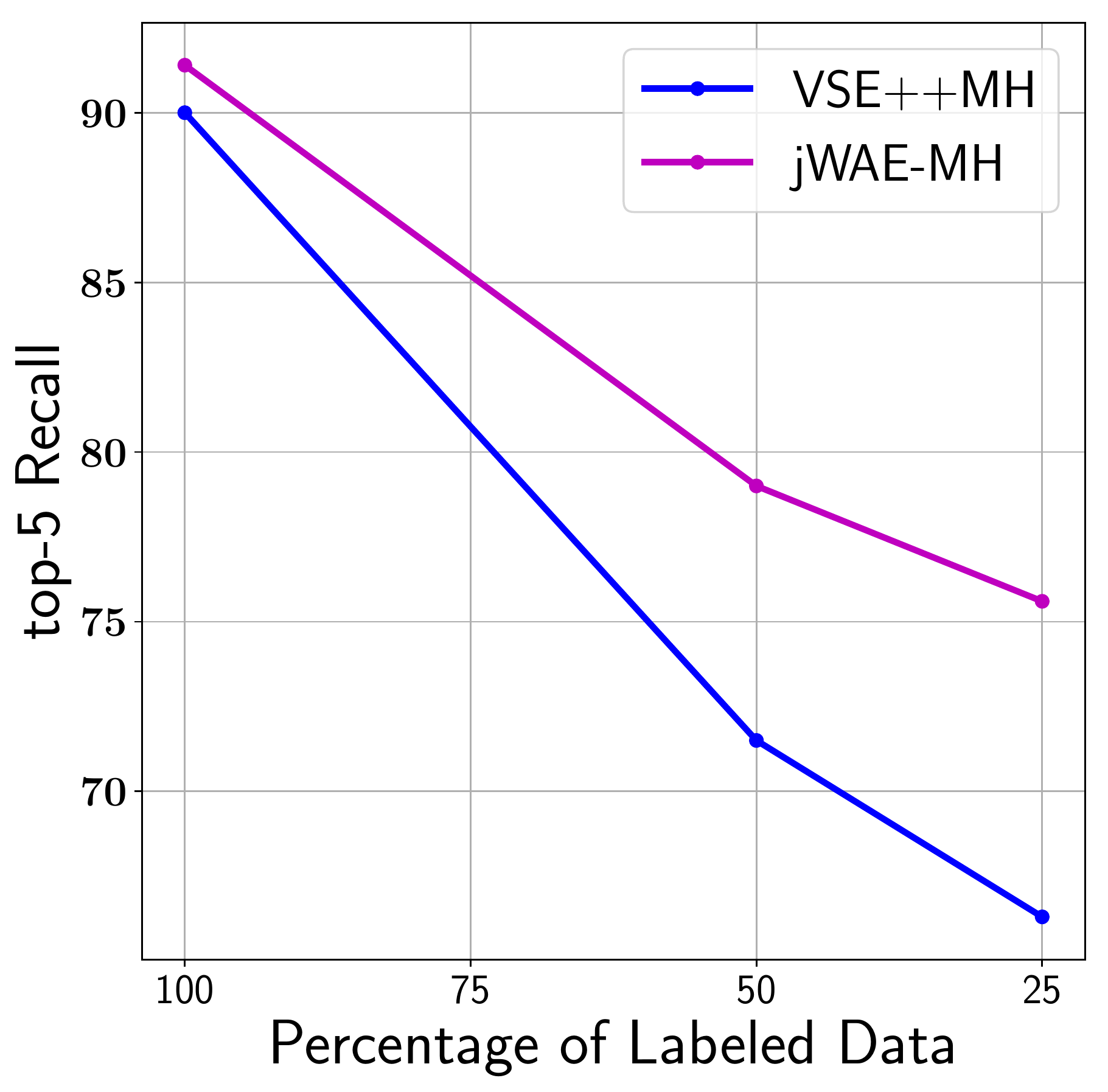}
\end{subfigure}%
\caption{Comparison of a standard supervised loss (MH) and the proposed semi-supervised jWAE under limited supervison: \emph{(left)} Flickr30k; \emph{(right)} COCO.}\label{fig:limitsup1}
\end{figure}

To show the effectiveness of our jWAE in minimally supervised settings, we perform additional experiments with limited labeled data in the training set, specifically with 25\% and 50\% of the labeled data.
If applicable, the rest of the data is included in an unpaired fashion for semi-/unsupervised learning.
We observe that compared to  standard supervised methods for learning joint embeddings of images and text \cite{FaghriFKF17},  
%The top-5 recall decays towards zero on reducing the number of supervised examples to 50\% (\cf \cref{fig:limitsup1}).
%In contrast, 
our semi-supervised approach performs better under different levels of supervision, highlighting the benefit of the semantic continuity from the joint Gaussian regularization. We observe that the accuracy gap widens as we decrease the supervision level, \eg from $7.4\%$ to $10.2\%$ points as we decrease the supervision from $50\%$ to $25\%$ on Flickr30k (\cf \cref{fig:limitsup1}).

\cref{table:limit_scan} shows the effect of training with limited supervision using region-based features and attention. Our variant of SCAN \cite{LeeCHHH18}, \ie~SCAN+jWAE-MH, outperforms SCAN by $3.4\%$ and $1.3\%$ on COCO and Flickr30k, respectively. 
Since the basic SCAN is trained with 36 image regions per image, which effectively increases the number of training points, SCAN performs reasonably well even at $25\%$ supervision. 
Adding our jWAE on top increases the robustness to limited supervision further.
Moreover, further decreasing the supervision to $5\%$, the top-1 recall of SCAN drops to $42.0\%$ while for SCAN+jWAE-MH the top-1 recall remains high at $54.3\%$ on the COCO dataset. 
This highlights the improved generalization performance of our jWAE framework at different supervision levels. 

\begin{figure}[t!]
\begin{subfigure}{.5\linewidth}
  \includegraphics[width=\textwidth]{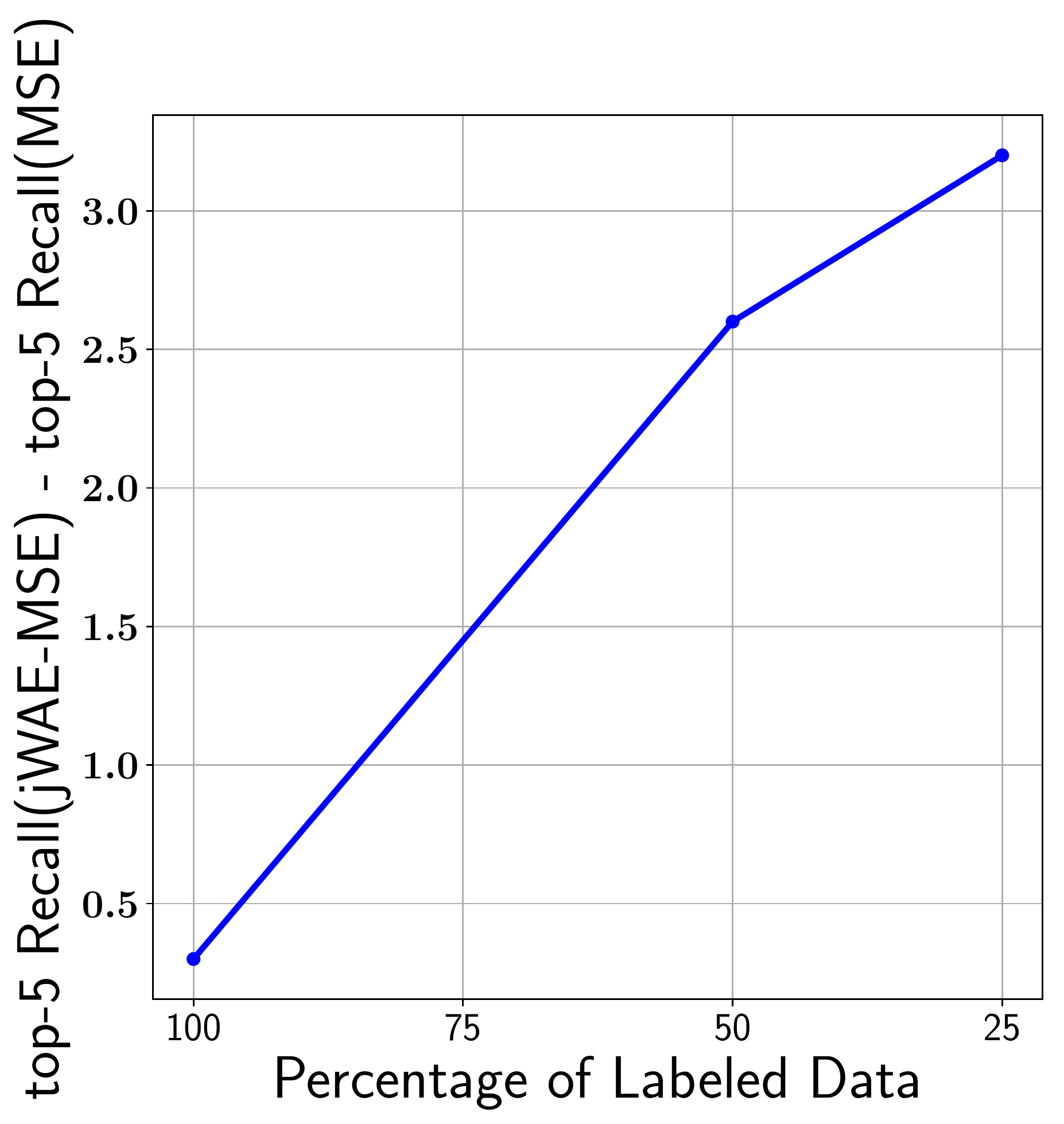}\end{subfigure}%
\begin{subfigure}{.5\linewidth}
  \includegraphics[width=\textwidth]{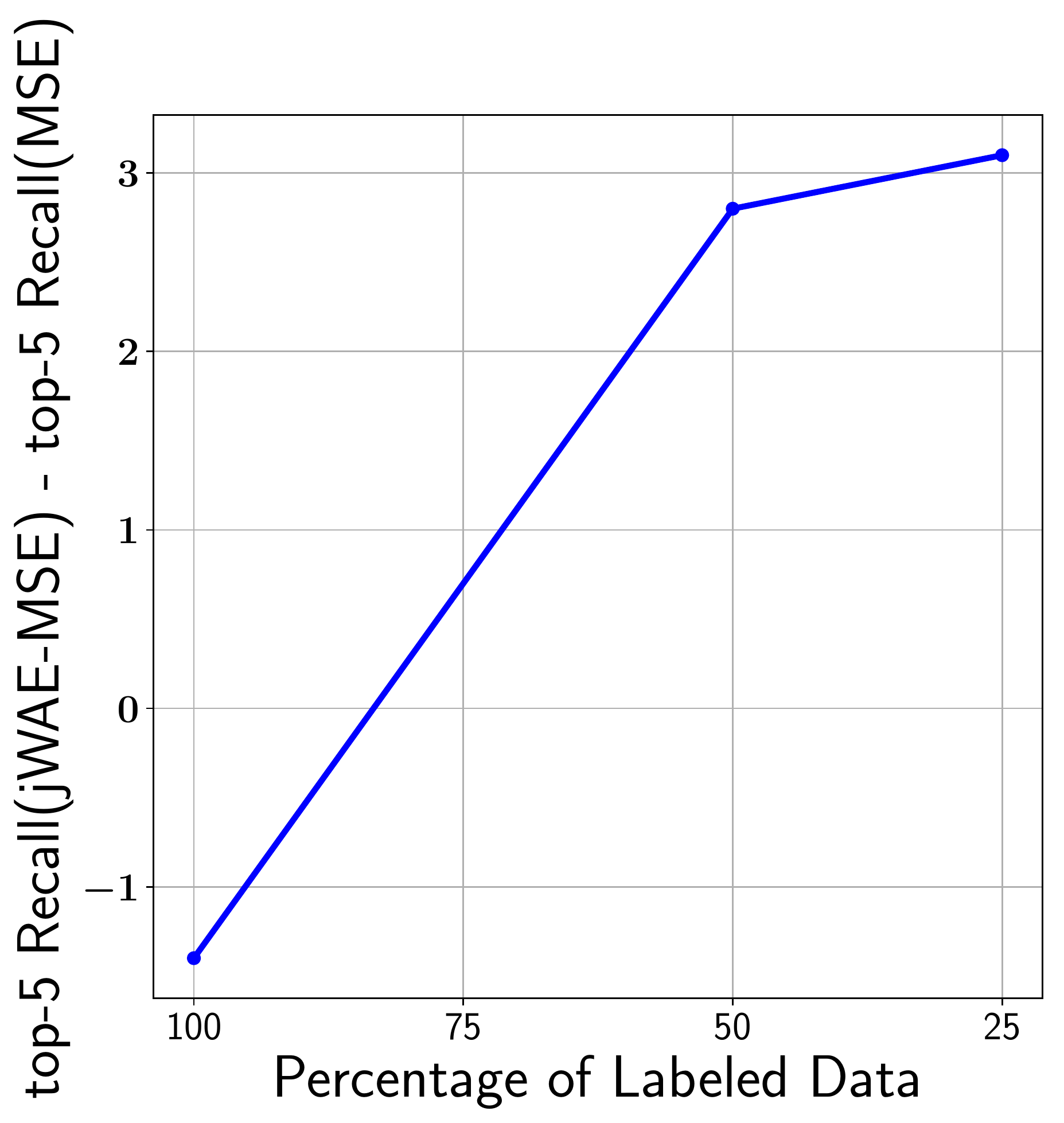}
\end{subfigure}%
\caption{Percentage gain in top-5 recall with the proposed jWAE-MSE over the MSE baseline under limited supervision with 50\% and 25\% of the labeled data: \emph{(left)} Flickr30k; \emph{(right)} COCO.}
 \label{fig:limitsup}
\end{figure}

We also compare against other methods that rely on additional unsupervised information, particularly the unsupervised autoencoder reconstruction loss combined with supervised MSE-based alignment, but without the Gaussian regularization (MSE). % We observe similar trends with respect to recall on . %As shown in Fig.~\ref{fig:limitsup},
The top-5 recall of our jWAE-MSE method is higher compared to the semi-supervised MSE baseline by 2.6\% and 3.2\% points on average for Flickr30k at 50\% and 25\% supervision, respectively.
Similarly for the COCO dataset, the top-5 recall is 2.8\% and 3.1\% points higher compared to MSE at 50\% and 25\% supervision, respectively, as shown in \cref{fig:limitsup}.
The difference in recall at a particular supervision level between our method and the baseline increases with decreasing labeled data.
This shows that encouraging continuity in the latent spaces helps to better align similar encoded representations and can be harnessed in learning cross-modal concepts even when supervised information is limited.

\begin{table}[t]
  \centering
  \footnotesize
  \begin{tabularx}{\columnwidth}{@{}XS[table-format=2.1]S[table-format=2.1]S[table-format=2.1]S[table-format=2.1]@{}}
    \toprule
    & \multicolumn{2}{c}{COCO} &\multicolumn{2}{c@{}}{Flickr30k}\\
    \cmidrule(lr){2-3}\cmidrule(l){4-5}
    Method(ResNet+GRU) (label \%) &  \multicolumn{1}{c}{R@1} &\multicolumn{1}{c}{R@5}&\multicolumn{1}{c}{R@1} &\multicolumn{1}{c@{}}{R@5} \\
    \midrule
    SCAN \cite{LeeCHHH18} (25\%) & 55.9 & 86.5 &45.9  & 72.6 \\
    SCAN \cite{LeeCHHH18} (5\%) & 42.0 & 84.3 &{--}  & {--} \\
   % VAE-MH  &57.8 &86.9 &46.3 & 80.3 \\
    SCAN+jWAE-MH (25\%) & 59.3 & 88.3 &  47.2 &  75.0  \\
    
   % VAE-MH  &57.8 &86.9 &46.3 & 80.3 \\
    SCAN+jWAE-MH (5\%) & 54.3 & 87.0 &  {--} &  {--}  \\
    \bottomrule
  \end{tabularx}
  \caption{Comparison of SCAN and SCAN-jWAE under limited supervision.
  %\textsuperscript{3}\footnotesize{Pre-trained model at \url{https://github.com/lwwang/Two_branch_network}} }
  }
  \label{table:limit_scan}
 \end{table}

\begin{table}%[t!]
\centering
\scriptsize
\begin{tabularx}{\linewidth}{@{}m{0.25\linewidth}m{0.3\linewidth}m{0.3\linewidth}@{}}
\toprule
\bfseries Text & \bfseries Ground truth image & \bfseries Retrieved image with jWAE-MSE  \\
\midrule
A man using his laptop computer while a cat sits on his lap. & \includegraphics[height=0.75\linewidth,width=0.9\linewidth]{2210_gt} & \includegraphics[height=0.75\linewidth,width=0.9\linewidth]{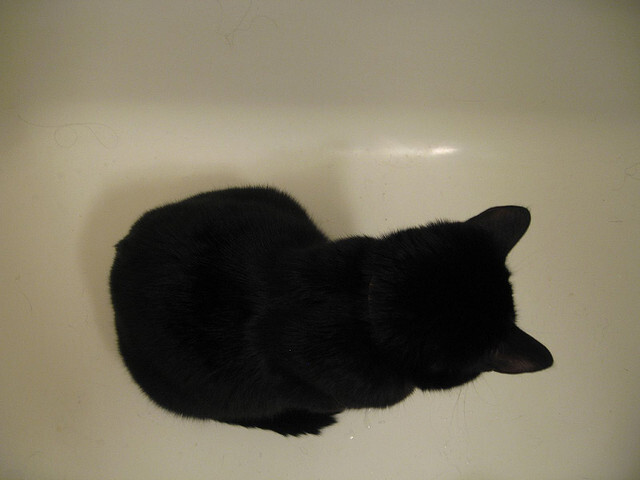}\\
A pizza and grapes sit on a tray next to a drink. & \includegraphics[height=0.75\linewidth,width=0.9\linewidth]{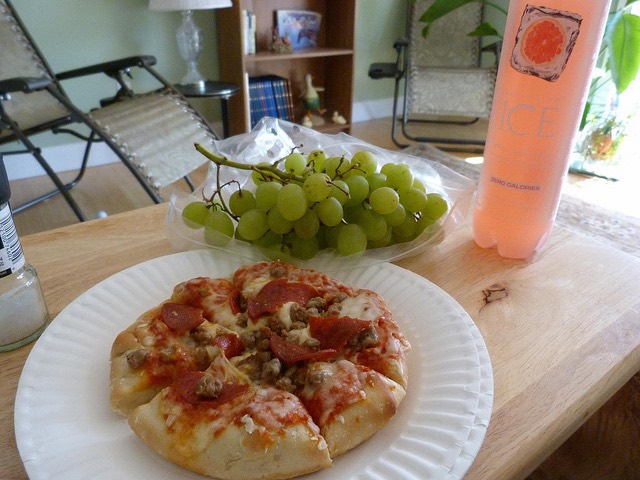} &  \includegraphics[height=0.75\linewidth,width=0.9\linewidth]{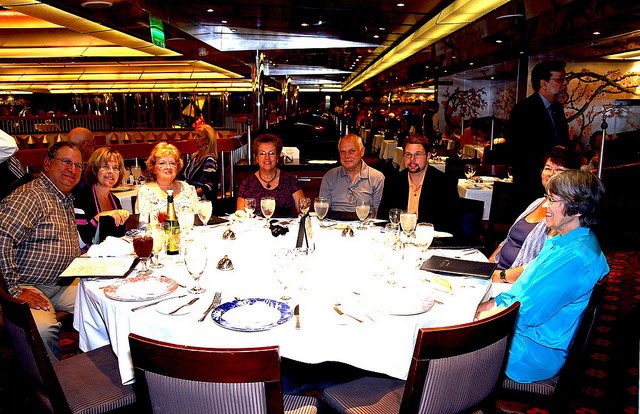} \\
A bath room with a toilet a sink and a mirror. & \includegraphics[height=0.75\linewidth,width=0.9\linewidth]{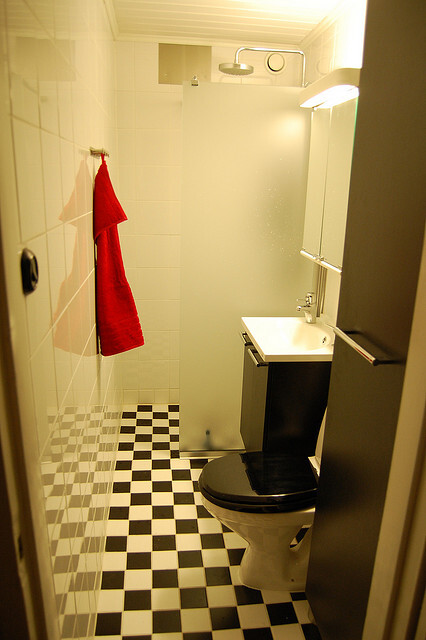} &  \includegraphics[height=0.75\linewidth,width=0.9\linewidth]{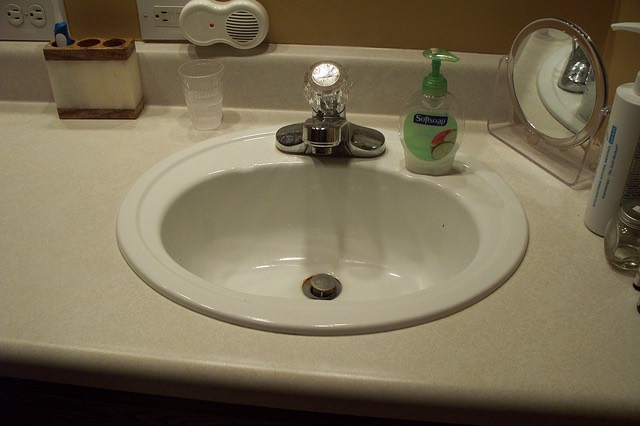} \\
A kitchen with a stove, oven, and refrigerator. & \includegraphics[height=0.75\linewidth,width=0.9\linewidth]{4678_gt} &  \includegraphics[height=0.75\linewidth,width=0.9\linewidth]{4678_m} \\
A slice of chocolate cake with dark chocolate icing. & \includegraphics[height=0.75\linewidth,width=0.9\linewidth]{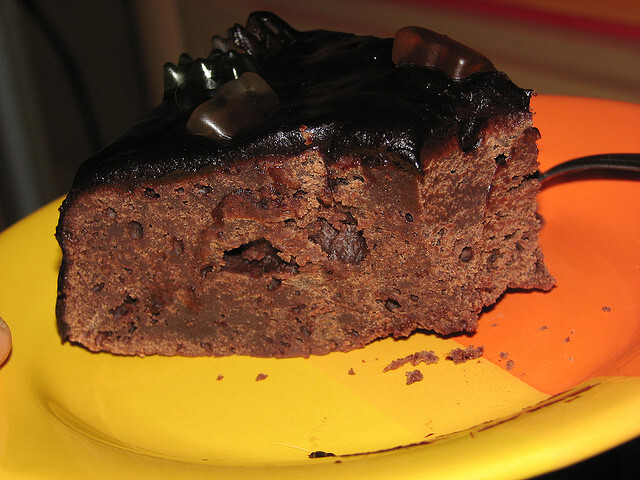} &  \includegraphics[height=0.75\linewidth,width=0.9\linewidth]{3379_m} \\
a no skate boarders sign on the side of the road. & \includegraphics[height=0.75\linewidth,width=0.9\linewidth]{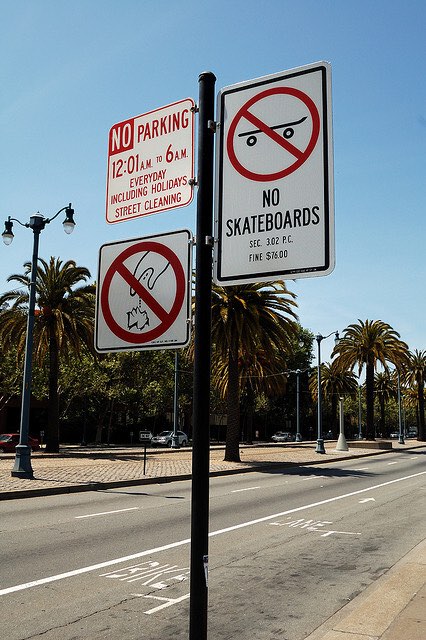} &  \includegraphics[height=0.75\linewidth,width=0.9\linewidth]{41_m} \\
\bottomrule
%\includegraphics[width= \linewidth, height=0.8\linewidth]{2406_gt} & \small\begin{itemize} \item Two boys sit shirtless on a trampoline.\item Two children are sitting shirtless on a trampoline . \itemA boy jumping on a trampoline . \item A blond girl and two American flags. \end{itemize}  \small\begin{itemize} \item Women play lacrosse.\item Two girls volleyball teams competing in a match. \itemTwo female soccer players compete for the ball. \item A crowd watching a soccer match. \end{itemize}\\

%\hline
\end{tabularx}
\caption{Given a caption, examples of images retrieved by our method that did not match the ground truth.
While they are clearly incorrect, they bear a semantic relation to the ground truth.}\label{table:sup_t2i}\end{table}

\begin{table}[t!]
\centering
\scriptsize
\begin{tabularx}{\linewidth}{@{}lX@{}}
\toprule
\bfseries Image & \bfseries Retrieved text with jWAE-MSE \\
\midrule
\raisebox{-\totalheight}{\includegraphics[width=0.3\linewidth, height=0.2\linewidth]{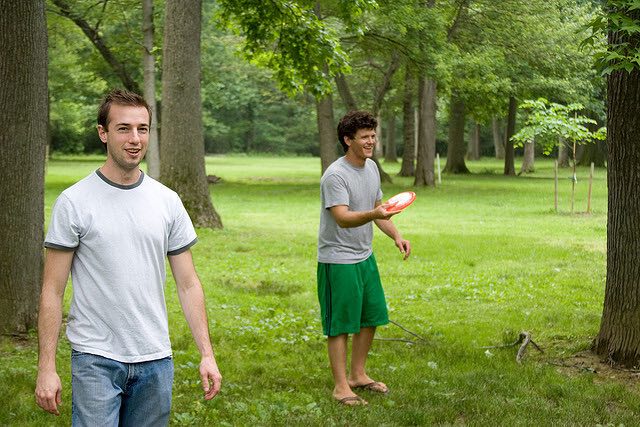}} & \begin{itemize}[nosep,leftmargin=*] \item A couple of men that are walking around on some grass. \item Two men are playing with a toy in a wooded area. \item  A couple of men standing on top of a lush green forest.  
\item{\bfseries A group of young men and women sitting at a table.} \end{itemize}\\

\raisebox{-\totalheight}{\includegraphics[width=0.3\linewidth, height=0.2\linewidth]{1160_gt}} & \begin{itemize}[nosep,leftmargin=*] \item Painting of oranges, a bowl, candle, and a pitcher. \item \textbf{Four boats with people  carrying lots of bananas and other foods.} \item \textbf{A couple of people in boats with food.}  \item  Painting of a table with fruit on top of it.\end{itemize}\\

\raisebox{-\totalheight}{\includegraphics[width= 0.3\linewidth, height=0.2\linewidth]{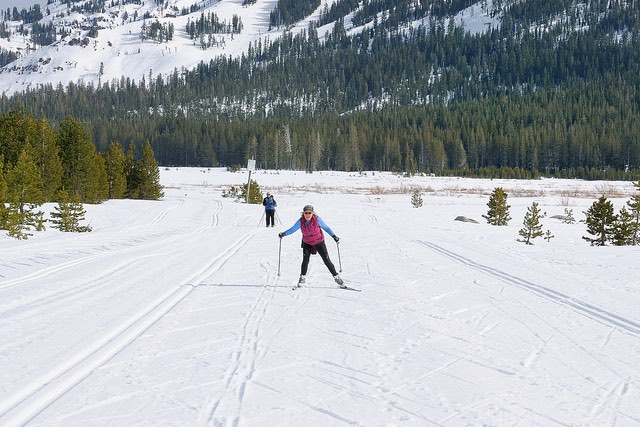}} & \begin{itemize}[nosep,leftmargin=*] \item A man riding skis down a snow covered slope. \item A person on skis going down a snow covered hill.\item A person turning while skiing down a snowy hill. \item A lady snow skiing on flat ground \end{itemize}\\

\raisebox{-\totalheight}{\includegraphics[width=0.3\linewidth, height=0.2\linewidth]{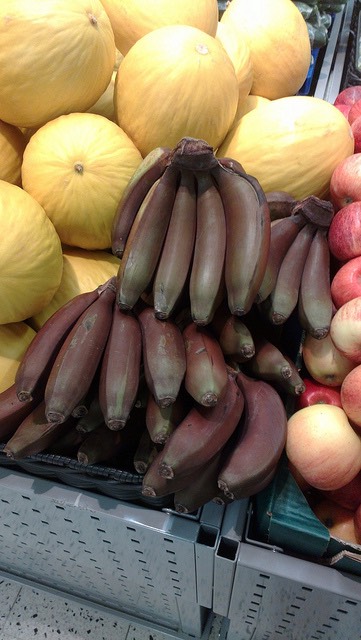}} & \begin{itemize}[nosep,leftmargin=*]  \item Some purple bananas and other fruits are together. \item Some purple bananas sitting between apples and some squash. \item A pile of black bananas and other fruit. \item Assorted fruit on display at a fruit market.\end{itemize}\\

\raisebox{-\totalheight}{\includegraphics[width=0.3\linewidth, height=0.2\linewidth]{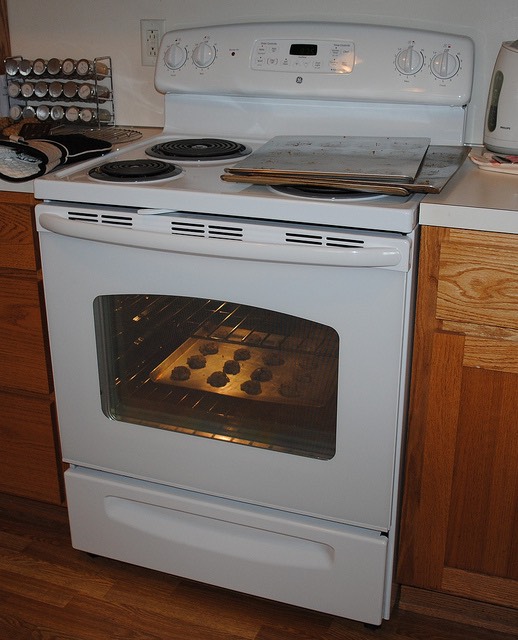}} & \begin{itemize}[nosep,leftmargin=*]  \item A kitchen with cookies in the oven baking. \item A white oven with cookies being baked inside. \item an oven with a pan of cookies baking inside it. \item The cookies are inside an oven in the kitchen.\end{itemize}\\

\raisebox{-\totalheight}{\includegraphics[width=0.3\linewidth, height=0.2\linewidth]{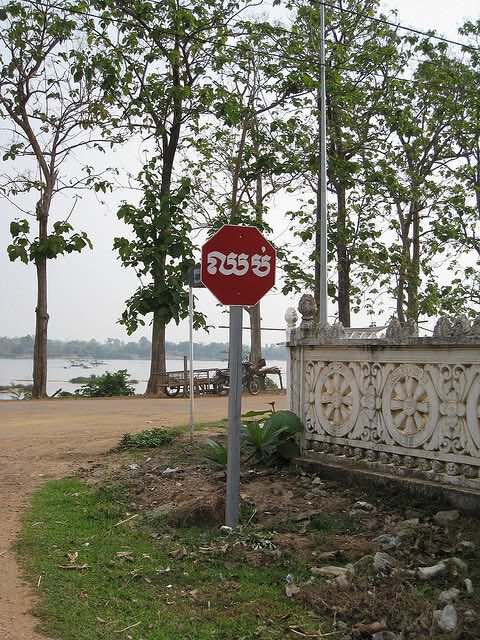}} & \begin{itemize}[nosep,leftmargin=*]  \item Foreign stop sign, possibly in Sanskrit or Cambodian script, with nice tree and water background, off-white property wall typical of India or southeast Asia. \item A stop sign in a foreign language by a body of water. \item A stop sign posted in a foreign language. \item A stop sign has a language that is not English.\end{itemize}\\
\bottomrule
\end{tabularx}
%}
\caption{Examples of the top-4 captions retrieved for a given image by the proposed jWAE-MSE. Captions that do not match the ground truth are shown in bold.}\label{table:sup_i2t}
\end{table}

\section*{Appendix B. Network Architecture}
\subsection*{Cross-modal retrieval}

\paragraph{Image pipeline.}
The input representations obtained from VGG-19 or ResNet-152 are fed into our joint Wasserstein autoencoder.
The image encoder takes 4096 inputs (2048 for ResNet-152), which are fully connected to a hidden layer of 2048 nodes.
The encoder outputs into a $d$-dimensional latent space. 
The decoder is symmetric to the encoder.
We apply ReLU nonlinearities after the hidden layer in both encoder and decoder.

\myparagraph{Text pipeline.}
We use a similar network with one hidden layer for the text encoder.
For 300-dimensional sentence inputs, the hidden encoder layer is of dimensionality 1024, and the latent layer has 256 dimensions. 
The same latent dimensionality is chosen for the image pipeline.
The decoder network is again symmetric to the encoder.
We apply ReLU nonlinearities after the hidden layer in the encoder and the decoder.
When using high-dimensional HGLMM features, we set the hidden layer size to 3000 and use a 512 (1024)-dimensional latent space for jWAE-MSE (jWAE-MH).

For GRU, the dimension of word embeddings is 300. Starting with one-hot coded vectors, the GRU encoder consists of an embedding layer, a bidirectional GRU layer, and a fully connected layer with dimensionality 1800.
The GRU decoder has a fully connected layer and a uni-directional GRU layer. Word dropout regulates the dependence of the prediction of a word in a sentence on the preceding ground truth sequence. To obtain a latent encoding that is representative of the sentence, we set the word dropout rate to 0.2 in the decoder.

\myparagraph{Discriminator network.}
We use a three layer discriminator network on the latent spaces of each pipeline. 
The sizes of the layers are chosen as 256, 256, and 2, respectively.
We use leaky ReLU activations between all the layers.
%The true labels for output of the discriminator are $0$ for true distribution and  $1$ for the competing distribution. 

\begin{figure*}[t]
  \centering
    \includegraphics[width=0.70\linewidth]{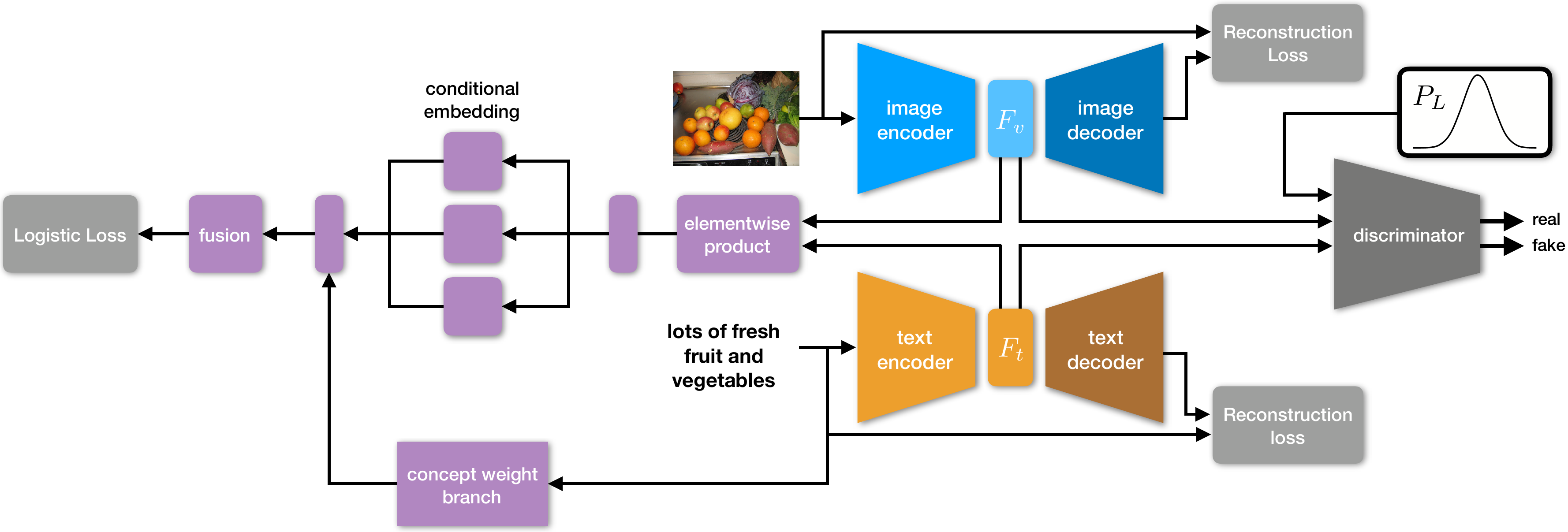}
      \caption{Architecture of CITE \cite{Plummer18} with our joint Wasserstein autoencoder.}
      \label{fig:archi_cite}
\end{figure*}

\subsection*{Phrase localization}
The encoder of the image pipeline passes the 4096-dimensional inputs through two fully connected layers, with 2048 hidden nodes and a 512-dimensional latent space. 
For the text pipeline, the 6000-dimensional input features are also passed through two fully connected layers, again with 2048 hidden nodes and 512 latent dimensions.
The decoders are symmetric to the encoders. We apply ReLU nonlinearities after the hidden layer in the encoder and the decoder.
For the discriminator network, we set the size of the three fully connected layers to 384, 256, and 2, respectively. Leaky ReLU nonlinearities are applied to the outputs of first and second layer. The last layer is a linear layer.
In \cref{fig:archi_cite}, we show CITE \cite{Plummer18} built upon our jWAE framework. The inputs to the image pipeline are 4101 dimensional with 4096-dimensional VGG-19 features and 5 spatial location features. Gaussian constraints are applied to the features from image and text pipeline before applying an elementwise product.

\section*{Appendix C. Cross-Modal Retrieval Results}
We provide some  examples for the cases where our method does not retrieve the correct image for text-to-image retrieval (\cf \cref{table:sup_t2i}). We observe that the retrieved images have similar concepts as the ground truth image. 
Without requiring any supervised information on image-image similarity, our jWAE is able to align similar concepts in images. 
This also shows that apart from the ground truth captions of the images, a caption can explain many images in the dataset.
In \cref{table:sup_i2t} we include results for image-to-text retrieval for jWAE-MSE showing that our method is able to recover relevant captions for a given image.

\myparagraph{Generalization across datasets.} In \cref{table:sup_crosst2i} and \cref{table:sup_crossi2t}, we show additional examples for a model trained on COCO and tested on the Flickr30k dataset.
We observe that our jWAE retrieves semantically more relevant images (texts) for a given text (image) compared to the competing method.
This shows that for good generalization performance of the embeddings, semantic continuity in the latent representations is desirable.

\begin{table}[t!]
\centering
\scriptsize
\begin{tabularx}{\linewidth}{@{}m{0.25\linewidth}m{0.33\linewidth}m{0.33\linewidth}@{}}
\toprule
\bfseries Text & \bfseries Matched images with supervised method \cite{wang2018learning} & \bfseries Retrieved image with jWAE-MSE  \\
\midrule
Two race cars are going down a racetrack bend. & \includegraphics[width=0.9\linewidth,height=0.8\linewidth]{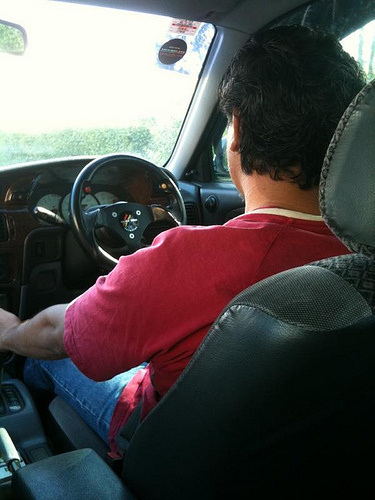} & \includegraphics[width=\linewidth,height=0.8\linewidth]{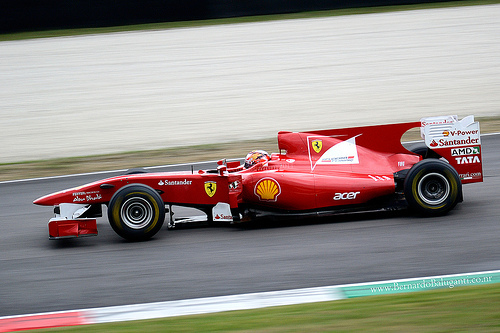}\\
Two women playing beach volleyball are jumping in the air for the ball. & \includegraphics[width=0.9\linewidth,height=0.8\linewidth]{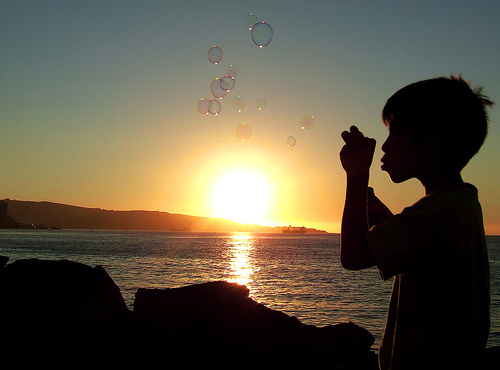} &  \includegraphics[width=\linewidth,height=0.8\linewidth]{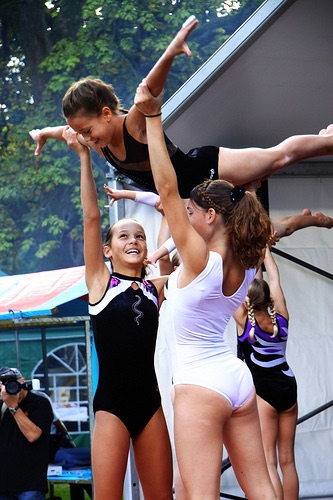} \\
Several cyclists make their way through a rutted field, as a cameraman films. & \includegraphics[width=0.9\linewidth, height=0.8\linewidth]{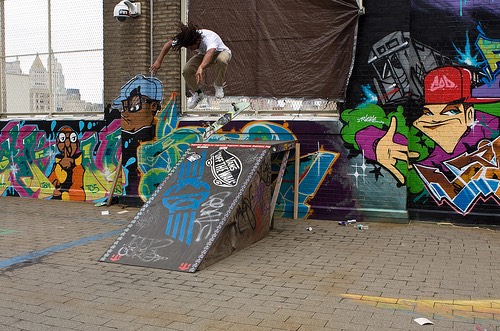} &  \includegraphics[width=\linewidth,height=0.8\linewidth]{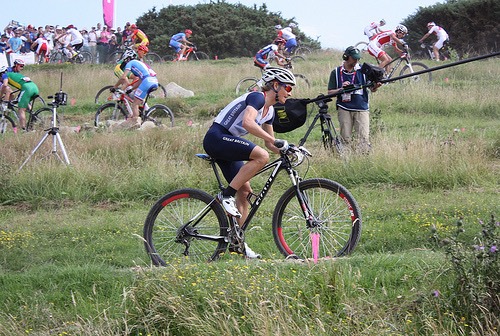} \\
A hockey goalie stands in front of the net at the ready. & \includegraphics[width=0.9\linewidth, height=0.8\linewidth]{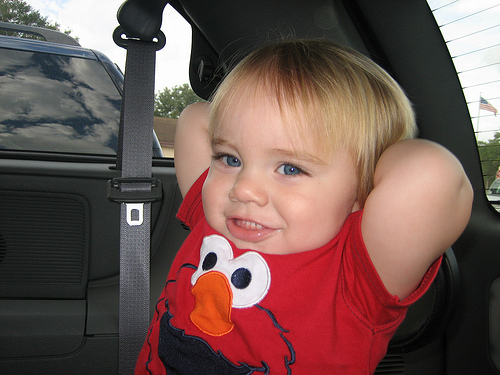} &  \includegraphics[width=\linewidth,height=0.8\linewidth]{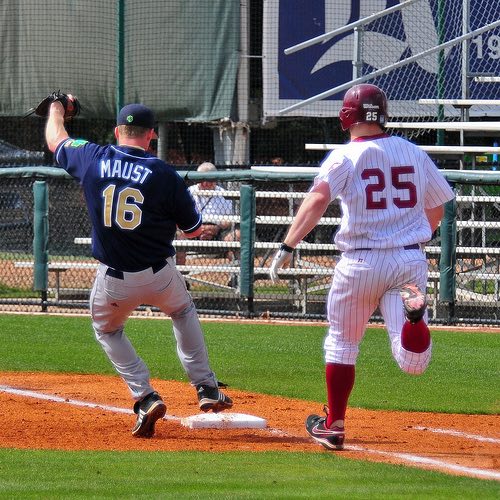} \\
A woman in white is holding a blue umbrella and walking in the rain. & \includegraphics[width=0.9\linewidth, height=0.8\linewidth]{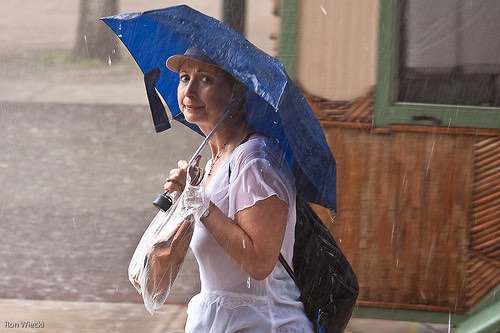} &  \includegraphics[width=\linewidth,height=0.8\linewidth]{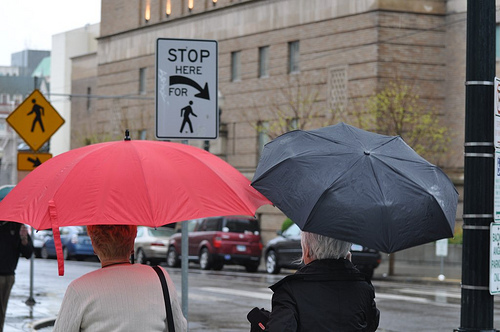} \\
Spray obscures the face and torso of the paddle wielding man in the red kayak. & \includegraphics[width=0.9\linewidth, height=0.8\linewidth]{2181_m} &  \includegraphics[width=\linewidth,height=0.8\linewidth]{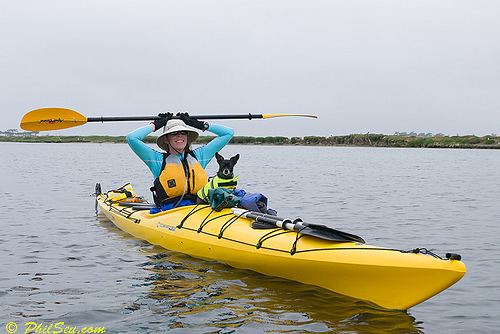} \\
 A line of people are biking down a road. & \includegraphics[width=0.9\linewidth, height=0.8\linewidth]{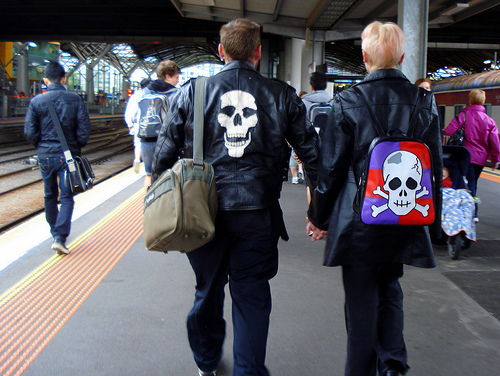} &  \includegraphics[width=\linewidth,height=0.8\linewidth]{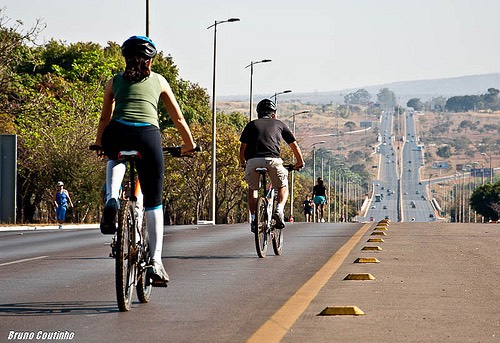} \\
\bottomrule
%\includegraphics[width= \linewidth, height=0.8\linewidth]{2406_gt} & \small\begin{itemize} \item Two boys sit shirtless on a trampoline.\item Two children are sitting shirtless on a trampoline . \itemA boy jumping on a trampoline . \item A blond girl and two American flags. \end{itemize}  \small\begin{itemize} \item Women play lacrosse.\item Two girls volleyball teams competing in a match. \itemTwo female soccer players compete for the ball. \item A crowd watching a soccer match. \end{itemize}\\
%\hline
\end{tabularx}
\caption{Examples of images retrieved for a given sentence on Flickr30k based on embeddings trained on COCO.}\label{table:sup_crosst2i}\end{table}

\begin{table}[t!]
\centering
\scriptsize
\begin{tabularx}{\linewidth}{@{}lXX@{}}
\toprule
\bfseries Image &
\bfseries Supervised approach \cite{wang2018learning} &
\bfseries Ours (jWAE-MSE)\\
\midrule
\raisebox{-\totalheight}{\includegraphics[width=0.3\linewidth]{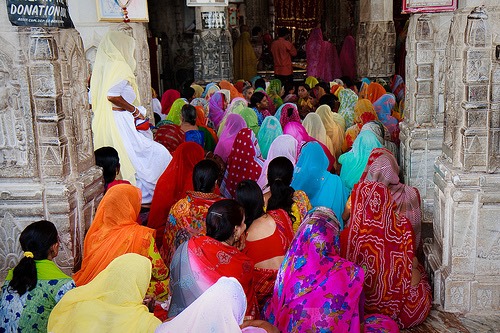}} & \begin{itemize}[nosep,leftmargin=*]\item People are at a costume party. \item A group of parents are sitting at an outdoor assembly. \item A group of friends working on a quilt. \item  A decorative plate with a piece of cake on it.
\end{itemize} & \begin{itemize}[nosep,leftmargin=*] \item A crowd of people in colorful dresses. \item Many ethnic people in multicolored robes gather for an event. \item A string instrument band and chorus singers rehearsing. \item A crowd of women wearing multiple different colors is entering a stone structure.
\end{itemize}\\[-6pt]
\midrule
\raisebox{-\totalheight}{\includegraphics[width=0.3\linewidth]{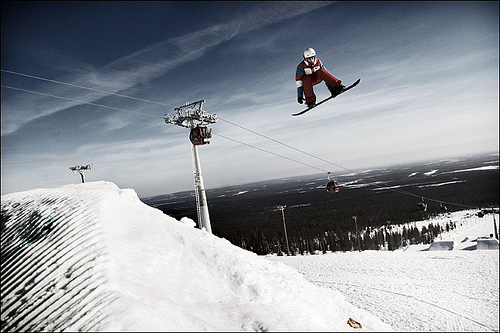}} & \begin{itemize}[nosep,leftmargin=*] \item A snowboarder soars through the air. \item  A stuffed bear with goggled on wearing snow skies. \item This is a skier getting some nice jumps in.  \item A young girl with her bike. \end{itemize} & \begin{itemize}[nosep,leftmargin=*] \item A snowboarder soars through the air. \item A snowboarder is doing a big jump and is in the air. \item Somebody is skiing down a mountain.  \item skier in red pants skiing down a slope. \end{itemize} \\[-6pt]
\midrule
\raisebox{-\totalheight}{\includegraphics[width=0.3\linewidth]{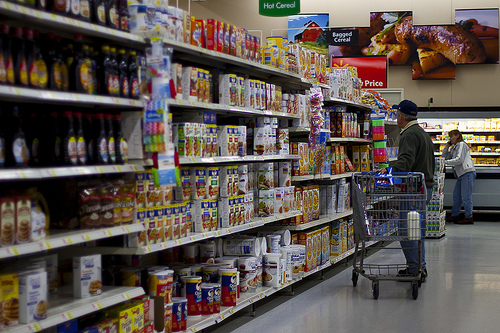}} & \begin{itemize}[nosep,leftmargin=*] \item People at a supermarket checkout. \item  A person wearing a hat made out of yellow bananas. \item Two men in orange construction hats are guiding a cart full of brick stones.  \item  A man with two kids looking at pictures on a camera. \end{itemize} & \begin{itemize}[nosep,leftmargin=*] \item A man is standing in the aisle of a grocery store and staring at the cereal selection. \item A customer at the checkout of a grocery store. \item A busy store full of people shopping at the store.  \item  A man with a shopping cart is studying the shelves in a supermarket aisle. \end{itemize} \\[-6pt]
\midrule
\raisebox{-\totalheight}{\includegraphics[width=0.3\linewidth]{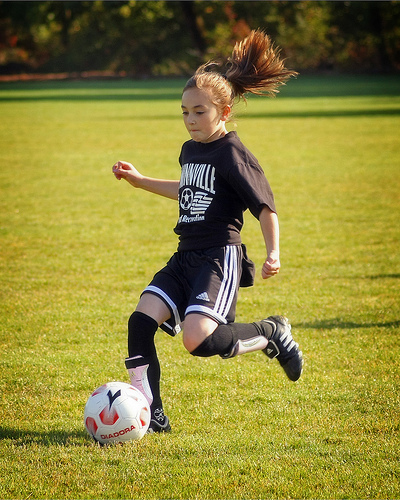}} & \begin{itemize}[nosep,leftmargin=*] \item A soccer player is kicking the ball. \item White and red striped bus riding through a city at night. \item A man in a blue apron in a kitchen.  \item  A young boy carrying a large soccer ball with a soccer feild in the background. \end{itemize} & \begin{itemize}[nosep,leftmargin=*] \item A soccer player is running while kicking a ball. \item A man in blue plays soccer. \item Two girls on separate teams fighting for a soccer ball.  \item  Two girls fight for the soccer ball while playing soccer on the grass.\end{itemize} \\[-6pt]
\midrule
\raisebox{-\totalheight}{\includegraphics[width=0.3\linewidth]{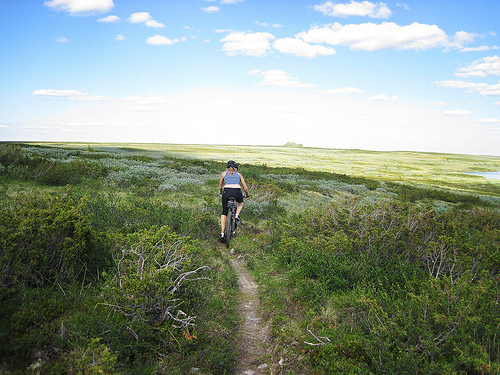}} & \begin{itemize}[nosep,leftmargin=*] \item A man jumping down a hill in a forested park. \item  A man jumps off a hill. \item A man sits on a field near a backpack.  \item  A zoo keeper tending to an elephant's mouth. \end{itemize} & \begin{itemize}[nosep,leftmargin=*] \item Mountain bike riders on a dirt trail. \item Woman on bicycle riding down dirt trail. \item Bicyclist riding on a dirt race course.  \item  Two women riding their bicycles along a dirt road. \end{itemize} \\[-6pt]
%\midrule
%\raisebox{-\totalheight}{\includegraphics[width=0.3\linewidth]{Images/sup-images/748_gt}} & \begin{itemize}[nosep,leftmargin=*] \item A man climbs up a mountain. \item  A person walks along rail road tracks in a wooded area. \item A climber scales a steep rock face.  \item  A man is sitting on top of a mountain. \end{itemize} & \begin{itemize}[nosep,leftmargin=*] \item A climber is scaling a mountain cliff. \item A male hiker sits perched on a high rock in the mountains. \item A man climbs up a mountain.  \item  A man wearing sunglasses is sitting on top of some jagged rocks with mountains in the background. \end{itemize} \\[-6pt]
\bottomrule
%\includegraphics[width= \linewidth, height=0.8\linewidth]{2406_gt} & \small\begin{itemize} \item Two boys sit shirtless on a trampoline.\item Two children are sitting shirtless on a trampoline . \itemA boy jumping on a trampoline . \item A blond girl and two American flags. \end{itemize}  \small\begin{itemize} \item Women play lacrosse.\item Two girls volleyball teams competing in a match. \itemTwo female soccer players compete for the ball. \item A crowd watching a soccer match. \end{itemize}\\
%\includegraphics[width=\linewidth, height=0.8\linewidth]{3862_gt} & \begin{itemize} \item Some purple bananas and other fruits are together. \item some purple bananas sitting between apples and some squash. \item A pile of black bananas and other fruit. \item Some purple bananas and other fruits are together \item Assorted fruit on display at a fruit market.\end{itemize}\\
%\hline
\end{tabularx}
\caption{Examples of the top-4 captions retrieved for a given image on Flickr30k based on embeddings trained on COCO.}\label{table:sup_crossi2t}\end{table}

\begin{table*}[t!]
\centering
\begin{tabularx}{\linewidth}{@{}m{0.3\linewidth}m{0.65\linewidth}@{}}
\toprule
\bfseries Images with bounding boxes & \bfseries Phrases (in red) in sentences \\
\midrule
\raisebox{-\totalheight}{\includegraphics[width=0.66\linewidth, height=0.528\linewidth]{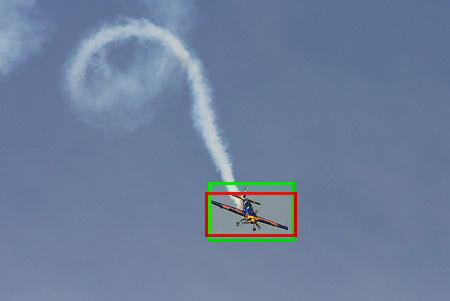}} & \textcolor{red}{A blue, red, and yellow airplane} is flying through the air.\\[-6pt]
\raisebox{-\totalheight}{\includegraphics[width=0.66\linewidth, height=0.528\linewidth]{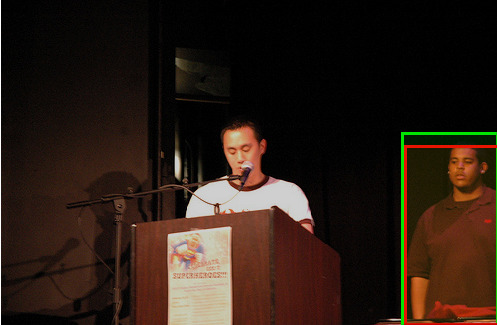}} & A young man in a t-shirt is speaking at a podium while \textcolor{red}{another young man} stands by.\\[-6pt]

\raisebox{-\totalheight}{\includegraphics[width=0.66\linewidth, height=0.528\linewidth]{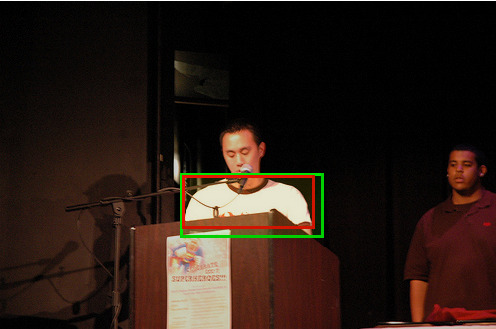}} & A young man in \textcolor{red}{a t-shirt} is speaking at a podium while another young man stands by. \\[-6pt]

\raisebox{-\totalheight}{\includegraphics[width=0.66\linewidth, height=0.528\linewidth]{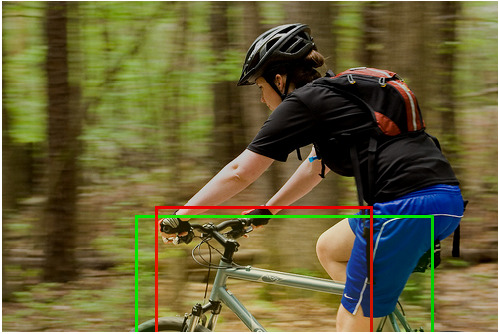}} & A woman rides her \textcolor{red}{bike} by some trees. \\[-6pt]

\raisebox{-\totalheight}{\includegraphics[width=0.66\linewidth, height=0.528\linewidth]{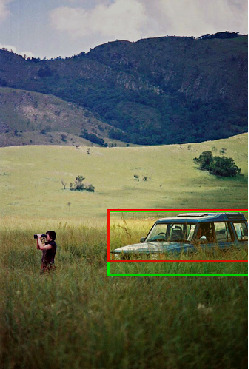}} & A person holding a piece of equipment up to her eyes is standing in a large meadow near a \textcolor{red}{blue vehicle}.\\[-6pt]

\raisebox{-\totalheight}{\includegraphics[width=0.66\linewidth, height=0.8\linewidth]{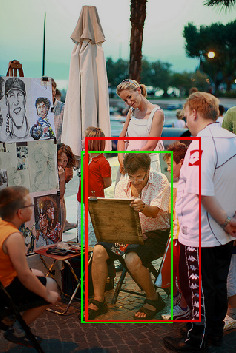}} & A crowd watches as a \textcolor{red}{woman} draws caricatures.\\
\bottomrule
\end{tabularx}
%}
\caption{Examples of the bounding boxes retrieved for a phrase in an image by the proposed jWAE. (\emph{green}) Ground truth bounding box; (\emph{red}) Retrieved bounding box.}\label{table:sup_phraseloc}
\end{table*}

\section*{Appendix D. Phrase Localization Results} In \cref{table:sup_phraseloc}, we include results of the bounding boxes retrieved by the proposed jWAE for the given phrase and image.
Our method is able to retrieve bounding boxes having high overlap with the ground truth bounding box.

\end{document}